\documentclass[shortAfour,sageh,times]{sagej}

\usepackage{moreverb,url}
\usepackage{balance}
\usepackage{caption}
\usepackage{subcaption}

\newtheorem{remark}{Remark}
\theoremstyle{remark}

\usepackage[usenames,dvipsnames,svgnames,table]{xcolor}
\definecolor{darkblue}{RGB}{0,0,175}

\usepackage[colorlinks,bookmarksopen,bookmarksnumbered,allcolors=darkblue]{hyperref}
\usepackage{url}
\usepackage{array}
\usepackage{booktabs}
\newcommand\BibTeX{{\rmfamily B\kern-.05em \textsc{i\kern-.025em b}\kern-.08em
T\kern-.1667em\lower.7ex\hbox{E}\kern-.125emX}}

\newboolean{include-notes}
\setboolean{include-notes}{true}
\newcommand{\ssnote}[1]{\ifthenelse{\boolean{include-notes}}%
 {\textcolor{red}{\textbf{SS: #1}}}{}}

 \newcommand{\cmnote}[1]{\ifthenelse{\boolean{include-notes}}%
 {\textcolor{OliveGreen}{\textbf{CM: #1}}}{}}

\newcommand{\jsnote}[1]{\ifthenelse{\boolean{include-notes}}%
 {\textcolor{blue}{\textbf{JS: #1}}}{}}
 
\newcommand{\rev}[1]{\textcolor{black}{#1}}

\def\volumeyear{2020}
\begin{document}

\runninghead{Lancaster {et~al.}}

\title{Electrostatic Brakes Enable Individual Joint Control of Underactuated, Highly Articulated Robots}

\author{Patrick Lancaster\affilnum{1}, Christoforos Mavrogiannis\affilnum{1}, Siddhartha Srinivasa\affilnum{1}, Joshua Smith\affilnum{1,2}}

\affiliation{\affilnum{1}Paul G. Allen School of Computer Science and Engineering, University of Washington, Seattle, USA\\
\affilnum{2}Department of Electrical and Computer Engineering, University of Washington, Seattle, USA}

\corrauth{Patrick Lancaster, planc509@cs.washington.edu}

\begin{abstract} 

Highly articulated organisms serve as blueprints for incredibly dexterous mechanisms, but building similarly capable robotic counterparts has been hindered by the difficulties of developing electromechanical actuators with both the high strength and compactness of biological muscle. We develop a stackable electrostatic brake that has comparable specific tension and weight to that of muscles and integrate it into a robotic joint. Compared to electromechanical motors, our brake-equipped joint is four times lighter and one thousand times more power efficient while exerting similar holding torques.  Our joint design enables a ten degree-of-freedom robot equipped with only one motor to manipulate multiple objects simultaneously. We also show that the use of brakes allows a two-fingered robot to perform in-hand re-positioning of an object 45\% more quickly and with 53\% lower positioning error than without brakes. Relative to fully actuated robots, our findings suggest that robots equipped with such electrostatic brakes will have lower weight, volume, and power consumption yet retain the ability to reach arbitrary joint configurations.  
\end{abstract}

\maketitle

\section{Introduction}
\begin{figure}[t]
\centering
\includegraphics[width=0.48\textwidth]{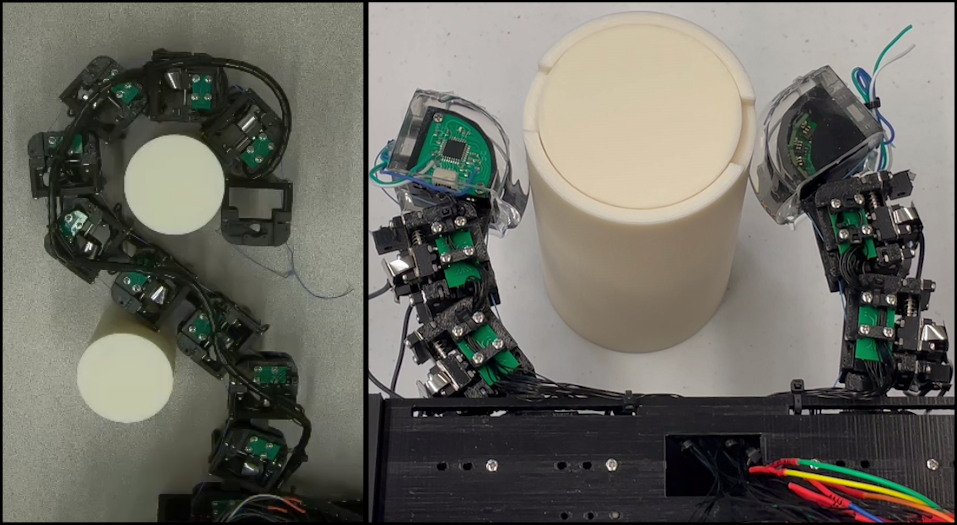}
\caption{Two underactuated, highly articulated robots built with our electrostatic brake equipped joint design. (Left) A ten degree-of-freedom serial chain robot with only a single motor is able to manipulate multiple objects simultaneously. (Right) A six degree-of-freedom robot hand with only two motors more quickly and precisely positions an object during in-hand manipulation when using embedded electrostatic brakes.
}
\label{fig:robots}
\end{figure}

Many organisms rely on the highly articulated nature of their bodies to perform vital functions. For example, human hands leverage twenty-one degrees-of-freedom (DoF) in order to nimbly manipulate objects~\citep{jones2006human}. Snake spinal columns composed of up to three hundred vertebrae facilitate versatile locomotion through their environments~\citep{ma1999analysis}. Lacking a rigid skeleton, octopus tentacles can use a seemingly unlimited number of degrees-of-freedom to both manipulate and locomote~\citep{mazzolai2007biorobotic}. While \rev{a large number of articulated joints} affords flexibility of movement, it is the networks of incredibly high strength to weight and volume ratio muscle that generate the forces necessary for such maneuvers~\citep{rospars2016force}.  Roboticists have attempted to solve highly dexterous tasks by building biologically inspired mechanisms with many degrees-of-freedom~\citep{tsai1995design,he2019underactuated,billard2019trends,transeth2009survey}, but their success has been limited by the inability of modern electromechanical actuators to simultaneously match both the strength and light weight of biological muscle. By designing an electrostatic brake with holding force that can be scaled to that of muscles yet remain similarly lightweight, we aim to build highly articulated, hybrid-actuated robot mechanisms with dexterity that approaches that of their biological counterparts (Fig. \ref{fig:robots}).

\begin{figure*}
  \includegraphics[width=\textwidth]{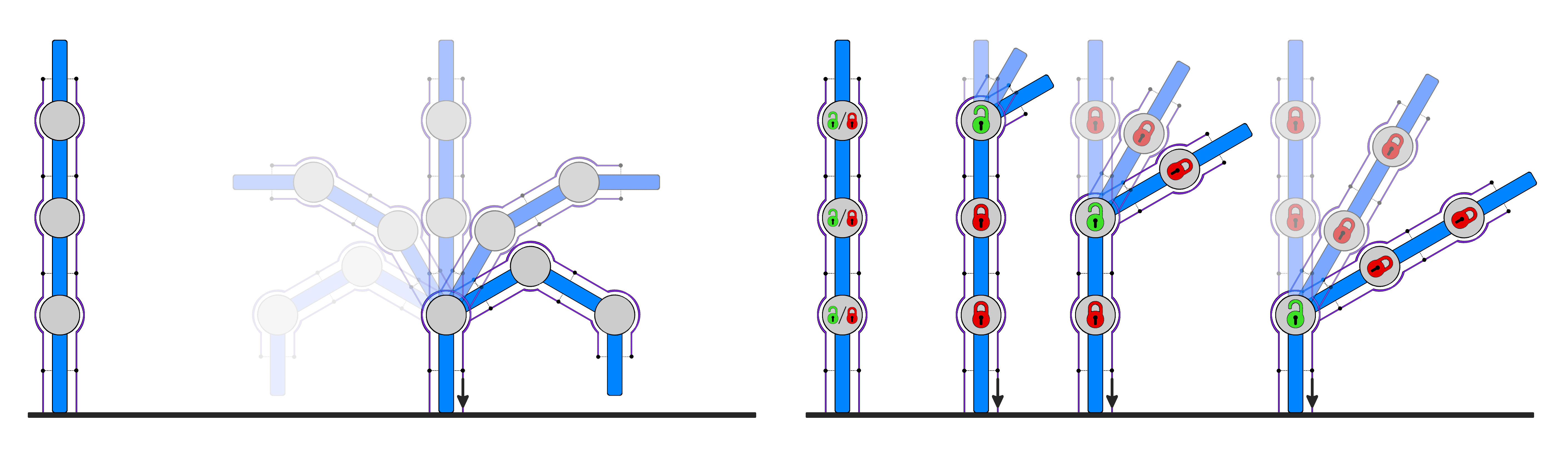}
  \caption{Installation of brakes into the joints of underactuated robots enables positional control of individual joints. Tension applied to the tendons exerts torque on all of the coupled joints but engaging a joint’s brake prevents it from moving. (Left) The range of motion of conventional underactuated systems is limited to the trajectories that occur when the tendons are pulled or released. (Right) By engaging any combination of brakes, a much wider range of trajectories can be achieved. }
  \label{fig:bata_whole}
\end{figure*}

When limited to conventional options for electronically controlled actuation, the most common architecture is to employ one or more motors per joint in order to achieve full actuation~\citep{tsai1995design,he2019underactuated}. However, applying this actuation scheme to mechanisms with a high number of degrees of freedom often violates volume, mass, power consumption, and cost constraints~\citep{billard2019trends,transeth2009survey}. To satisfy these constraints, roboticists tend to reduce the number of motors by coupling multiple joints together through the use of tendons (Fig.~\ref{fig:bata_whole}). However, this severely degrades the space of joint trajectories that the mechanism can execute because direct control of individual joints is no longer possible~\citep{ma2017yale,gosselin2008anthropomorphic,grioli2012adaptive,xu2012design}.

Positional control of each individual joint can be restored to such tendon-driven systems by installing a brake into each joint. This allows torque exerted on a joint by a motor-actuated tendon to be counteracted by engagement of the joint’s brake. Here, it is assumed that each brake can only be in one of two states (on or off), and that, when engaged, the applied braking torque entirely negates the torque applied by the tendon on the corresponding joint. Thus, by selectively choosing which brakes are engaged, joints that are connected to the same tendon can have their motion decoupled (Fig. \ref{fig:bata_whole}). \rev{Any subset of joints can be held stationary by engaging their corresponding brakes, while the remaining unbraked joints are free to move according to the torque exerted by the tendon (where we demonstrate a practical implementation of such actuation in Section \ref{sec:chain_desc})}. This actuation strategy can be applied to any kinematic structure in which the joints are coupled together by one or more tendons~\citep{jacobsen1989antagonistic,koganezawa1999mechanical}. Although the mechanism is still underactuated, its dexterity has been significantly improved because the restoration of control over individual joints allows it to reach arbitrary joint configurations.

\begin{figure}[t]
\centering
\includegraphics[width=0.5\textwidth]{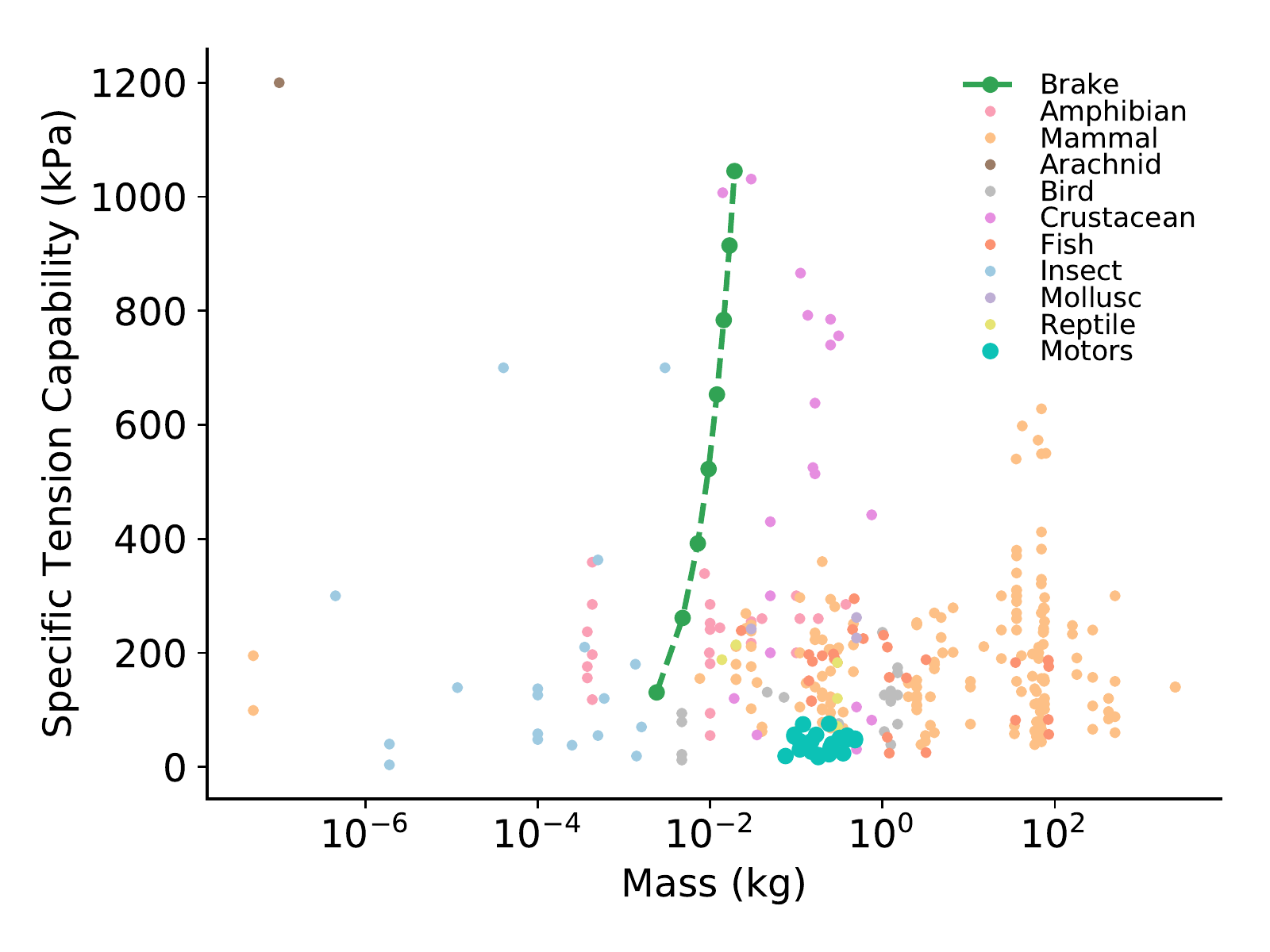}
\caption{ The expected frictional force per cross-sectional area (i.e. specific tension) of our electrostatic brakes is compared to that of measured biological muscle~\citep{rospars2016force} across multiple taxonomic groups. Braking force can be scaled by stacking sets of thin electrodes on top of each other (up to 8 stacked sets shown). Please see the beginning of Section \ref{sec:brake_performance} for brake parameter values.
}
\label{fig:muscle_compare}
\end{figure}

Electrostatic brakes offer a promising pathway towards the implementation of dexterous brake-aided systems. Although the muscles that power dexterous movement in biological organisms use different physical mechanisms for force generation than electrostatic brakes, the metric of specific tension (i.e., force per cross-sectional area) that has classically been used to measure muscle strength is also naturally applied to electrostatic brakes. Muscles typically have maximum specific tension exertion capability on the order of two hundred kilopascals across a wide range of taxonomic groups~\citep{rospars2016force}, but electrostatic brakes can achieve similar or higher specific tension while having a lower mass than most types of muscle (Fig. \ref{fig:muscle_compare}). This is particularly evident for brakes that stack two or more sets of electrodes on top of each other. Electromechanical motors are significantly heavier and, although they may be able to exert comparable amounts of raw force/torque, their relatively large cross-sectional areas result in worse specific tensions in comparison to electrostatic brakes and most types of muscle \rev{(see Remark~\ref{remark:motors-vs-brakes})}.


 \rev{Achieving the full strength of an electrostatic brake requires maximum conformance between the brake's electrodes.} Thick electrodes lack the compliance necessary to properly conform to one another. Thin electrodes will buckle if they are not tensioned along the direction of an applied force, or incidentally unzip if out-of-plane forces are applied at the electrodes' edges. Our insight is that we can optimize thin electrode conformance by employing a rack-and-pinion transmission to convert rotational motion of the joint into linear sliding between the electrodes. Controlling the motion of the electrodes to be along a single linear axis eliminates the out-of-plane forces that cause unzipping, and makes it easy to tension the electrodes along the direction of motion in order to prevent buckling. Carefully choosing the thickness and spacing of the electrodes allows this optimal conformance to be extended to multiple stacks of electrodes. By optimizing electrode conformance and stacking thin electrodes, we can achieve significant braking strength without increasing brake area or employing expensive, specialized dielectrics.

We transform this insight for optimal brake conformance into a novel joint design. With this joint design, we aim to demonstrate that hybrid motor-brake actuation enables robot operation in scenarios that not only require dexterity, but also place constraints on the robot's weight, size, and/or power consumption. To that end, we first measure the braking capabilities of this design in order to verify that it achieves the significant braking capability predicted by its theoretical, ideal-electrode-conformance model. \rev{Motivated by applications that constrain mechanism size, weight, and power consumption, we compare our joint design to robot joints driven by electromechanical motors and } find that our brake equipped joint exerts similar holding torques yet is significantly more compact, lightweight, and power efficient. We then build two different brake-equipped robots that demonstrate how our brake design significantly enhances the dexterity of underactuated robots by enabling control of individual joints. The first robot uses only a single motor and its brakes to actuate its ten degrees-of-freedom. This robot is able to manipulate multiple objects simultaneously, a task that a conventionally underactuated robot counterpart would not be able to complete. Finally, we construct a two-fingered robot hand with six degrees of freedom and two motors, and then measure its ability to perform an in-hand manipulation task in which the robot must move an object from one side of its workspace to the other. While the hand can complete the task without brakes, the use of brakes reduces both the execution time and final positioning error by approximately 50\%. A playlist of supplementary videos is available here: \url{https://www.youtube.com/playlist?list=PLgqhWpj1Z3vNHvt62tlPj-QQyJd2clADj}

\begin{remark}{\label{remark:motors-vs-brakes}\rev{It is important to note that unlike biological muscle and electromechanical motors that can independently exert force/torque, electrostatic brakes can only resist force/torque. The application of electrostatic brakes is most relevant to scenarios in which constraints on weight (and size and power as we discuss later) make it infeasible to fully actuate the mechanism with conventional electromechanical motors. In such regimes, hybrid motor-brake actuation, as illustrated in Fig. \ref{fig:bata_whole}, has the potential to satisfy these constraints while still enabling the robot to reach any valid joint configuration.} }\end{remark}

\section{Related Work}
There are a number of applications in which the need for lighter, more compact, power-efficient, and inexpensive actuation has prompted the use of electrostatic brakes. Such applications include communications equipment, tactile displays, virtual reality, and exoskeletons. \rev{While the design of our electrostatic brakes was most influenced by works from these previous fields that are somewhat tangential to robotics, we also describe} electrostatic force generation technologies (including but not limited to electrostatic braking) that have been applied to robotic grasping, locomotion, and modular robotics for completeness. 

\subsection{Electrostatic Brake Applications}

\citet{johnsen1923physical} perform some of the earliest investigations into electrostatic braking. They provide empirical observations of the effects of applied voltage and conductor separation distance on the resulting electrostatic force, and find that the amount of current required to achieve significant electrostatic forces is small (i.e. on the order of a few microamperes). Furthermore, they note that using a dielectric to separate the conductors of the brake with high voltage difference can lead to a buildup of charges in the dielectric that is detrimental to the observed braking effect. They instead separate the conductors with a rigid semi-conductive layer and model the effective separation distance between the conductors as the air gap(s) between the semi-conductive layer and the conductors. Such air gaps are the result of the conductive and semi-conductive layers' inability to perfectly conform to one another due to rigidity and imperfect surface smoothness. \citet{johnsen1923physical} also describe the zippering effect in which the points at which the conductors are closest act to pull nearby points closer to each other. Finally, they use electrostatic brakes to create multiple types of equipment for use in telecommunications, namely an electric relay, telegraph recorder, and a loud-speaking telephone.

More recently, electrostatic braking has been used to create various types of haptic interfaces. \citet{zhang2018electrostatic} use electrostatic brakes to control the pins of a 2.5D tactile display. Given a desired tactile pattern, a linear actuator extends in order to push all of the pins up and out of the base of the display. As the linear actuator contracts and allows the pins to fall, each individual pin's electrostatic brake will engage in order to form the overall desired tactile pattern. Each pin of the 4-by-2 array can withstand at least 28 gram-force of static loading force applied by the user. 

\rev{The design of our electrostatic brake is most similar to those of \citet{hinchet2018dextres} and \citet{diller2016lightweight} in which brake engagement prevents linear sliding between pairs of electrodes.} \citet{hinchet2018dextres} create a virtual reality glove that can restrict the movement of the thumb and forefinger via electrostatic brakes. Activating the brakes when grasping a virtual object provided the users with effective haptic feedback and improved grasp precision. \citet{diller2016lightweight} use electroadhesive clutches to design an ankle exoskeleton. Activation of each individual clutch causes a corresponding spring to engage with the exoskeleton. By activating different combinations of clutches, \citet{diller2016lightweight} are able to produce six different levels of stiffness throughout the user's walking gait. Their electroadhesive clutch achieves three times higher torque density while consuming two orders of magnitude less power per unit torque relative to other electronically controlled clutches. \rev{Rather than using the specialized insulators with large dielectric constants that \citet{hinchet2018dextres} and \citet{diller2016lightweight} employed to generate high braking forces, we instead stack multiple
thin electrodes on top of each other to achieve a compact brake design with inexpensive materials.}

\subsection{Electrostatic Technologies in Robotics}

The underlying principles of electrostatic braking have been applied to a number of different tasks in robotics. Dieletric elastomer actuators (DEA) consist of two conductors separated by a compressible dielectric~\citep{pelrine2000high}. The attractive electrostatic force between the conductors resulting from applying a large voltage across them produces significant strains in the dielectric. \citet{ji2019autonomous} integrate three DEAs into an insect sized robot. Each DEA actuates a leg at 450 hz in order to steer the robot and move at speeds of up to 3 cm/s. \citet{prahlad2008electroadhesive} use electroadhesive technology to create robots that can locomote along walls. Given a sequence of highly compliant electrodes embedded in the robot, adjacent electrodes are oppositely charged in order to induce reciprocal charges in the wall substrate. The resulting attractive electrostatic forces allow both tracked and biomemetic robots to climb walls of many different surface types. Similar electroadhesive methods have been applied to robot grasping, particularly to manipulate objects that are irregularly shaped or delicate~\citep{guo2018soft, savioli2014morphing,shintake2016versatile}. \citet{karagozler2007electrostatic} use electrostatic force generation as a latching mechanism for modular robots. They also explore inter-modular power transfer and communication via the capacitive coupling that the electrostatic latch provides.

\citet{aukes2014design} create an underactuated robotic gripper that uses electrostatic brakes to achieve hand configurations useful for grasping, as well as to increase the maximum pullout force of power grasps. Their work is the most similar to our own in that it uses electrostatic brakes to control the motion of the joints in an underactuated, tendon-driven mechanism. Although their brakes allowed the underactuated hand to reach discrete configurations that would not otherwise be possible, the use of large DC voltages to engage the brakes caused dielectric charge injection that makes engagement and disengagement of the brakes relatively slow. This limits the use of braking to discrete changes in hand configuration rather than continuous control. 

A number of works have shown that the effects of charge injection are mitigated by using AC voltages to actuate the brakes~\citep{hinchet2018dextres,hinchet2020high,zhang2018electrostatic, diller2016lightweight,karagozler2007electrostatic}, and we have created our own hardware to implement such a strategy. We also propose a novel brake-aided joint design; we use a rack and pinion transmission system to convert rotational joint motion into linear sliding between the electrodes, whereas the brake electrodes in \citet{aukes2014design} rotate around the axis of the joint. Unlike \citet{aukes2014design}, we demonstrate that our joint design's observed braking strength approximately matches that of the theoretical model, including when stacking multiple brake electrodes on top of each other. Furthermore, while \citet{aukes2014design} demonstrate electrostatic braking for robot grasping, we examine the use of brakes for in-hand manipulation. Finally, their robot gripper actuates at most three degrees of freedom per motor. In contrast, we demonstrate not only how electrostatic braking can enhance the dexterity of robot hands, but also how this technology can be scaled to robots with increasingly large joint to motor ratios. For the first time, we extend electrostatic braking to mechanisms with up to ten degrees of freedom per motor that can find use as robotic snakes or tentacles. 

\section{Electrostatic Force Generation}

An electrostatic brake can be formed by bringing two conductors (or generally two bodies with conductors attached to them) in contact with a dielectric that separates them. When voltage is applied across the conductors, the resulting attractive force induces a frictional force that resists tangential motion between the conductors. Thus, electrostatic brakes may oppose relative motion between two bodies by inducing frictional forces between them.


The maximum braking force is primarily determined by the area of overlap between the conductors, applied voltage, separation distance, dielectric permittivity, and the conductor-dielectric coefficient of friction. However, achieving a desired braking force is limited or hindered by a number of factors such as the voltage tolerance of control electronics, reduced effective area of overlap due to poor electrode conformance, and cost of high permittivity dielectrics. In the following subsections, we formalize the modeling of electrostatic brake strength, discuss the design considerations necessary for achieving a desired braking capability, and provide motivation for a brake design that facilitates the stacking of brakes.

\subsection{Parallel Plate Capacitor Model}
\label{sec:cap_model}

\begin{figure}[t]
\centering
\includegraphics[width=0.5\textwidth]{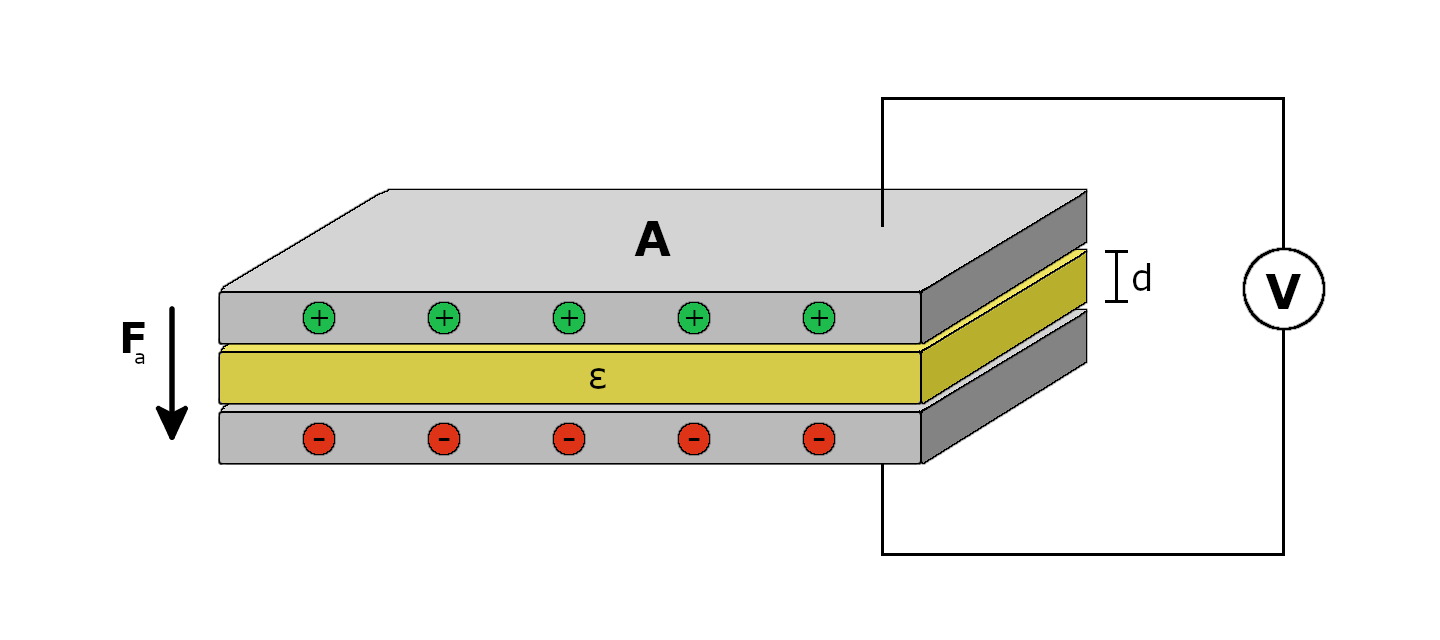}
\caption{ Electrostatic phenomena between two parallel conductors can induce a braking effect. In the absence of applied voltage, the two parallel conductors can exhibit planar sliding motion relative to each other. Once voltage is applied, an attractive force between the oppositely charged conductors induces a frictional force between them that prevents sliding.
}
\label{fig:brake_physics}
\end{figure}

The most basic brake structure consists of two conductive plates separated by an insulator. A controlled voltage applied to the brake creates an attractive force $F_a$ between the plates, which can be modelled as a parallel plate capacitor (Fig. \ref{fig:brake_physics}): 

\begin{equation}
    F_a = \frac{\epsilon A V^2}{2 d^2}\mbox{,}\label{eq:F2V}
\end{equation}

\noindent where $\epsilon$ is the permittivity of the dielectric, $A$ is the area of overlap, $V$ is the applied voltage, and $d$ is the distance between the plates. Along with the coefficient of friction between the plates and dielectric $\mu$, $F_a$ induces a frictional force $F_{max \:brake}$ that opposes sliding motion of the plates relative to each other:

\begin{equation}
    F_{max\:brake} = \mu F_a\mbox{.}
\end{equation}

\begin{figure}
     \centering
     \begin{subfigure}[b]{0.1\textwidth}
         \centering
         \includegraphics[width=\linewidth]{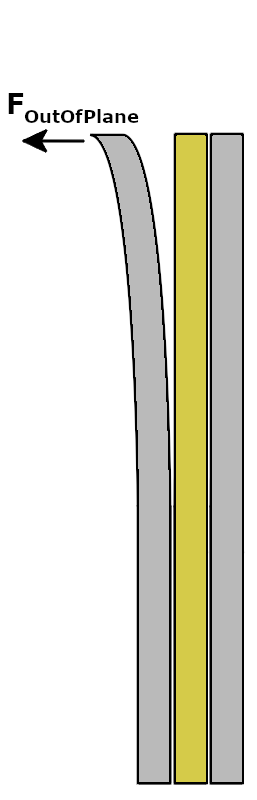}
     \end{subfigure}
     \hspace{0.15\textwidth}
     \begin{subfigure}[b]{0.1\textwidth}
         \centering
         \includegraphics[width=\textwidth]{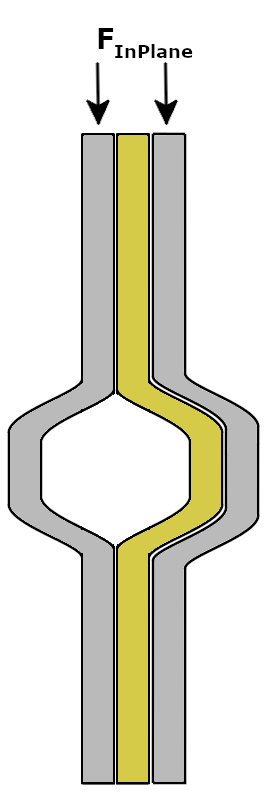}
     \end{subfigure}

        \caption{\rev{Poor conformance between electrodes reduces braking strength. (Left) Unzipping: Out-of-plane forces at the edges of overlap can cause the electrodes to peel away from each other. (Right) Buckling: Without proper tensioning of the ends of the electrodes, in-plane forces can cause compliant electrodes to collapse in on themselves.}}
        \label{fig:non_ideal}
\end{figure}

\begin{figure*}
  \includegraphics[width=\textwidth]{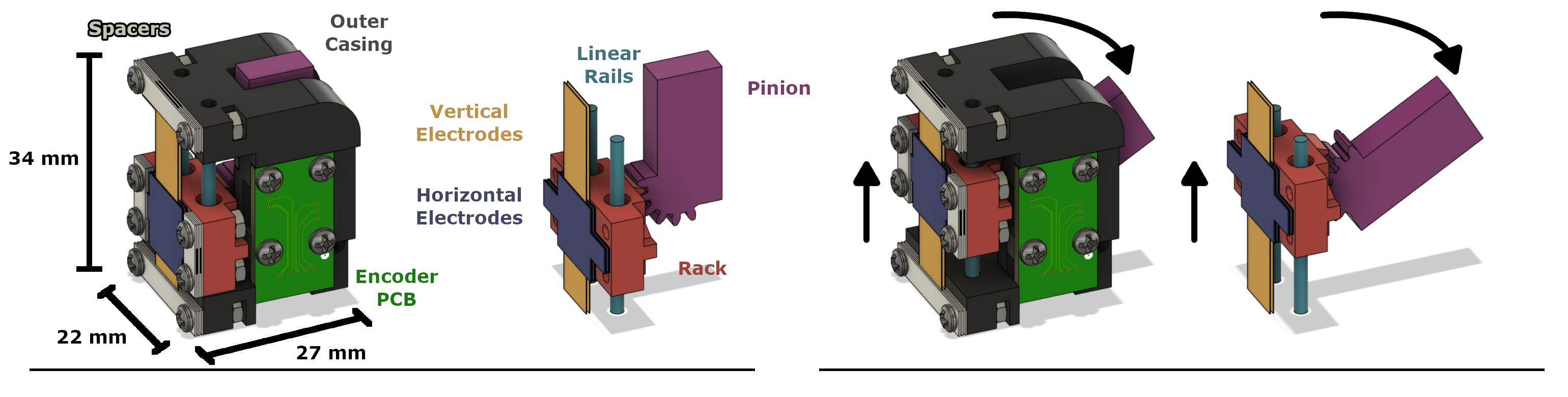}
  \caption{Rotational motion of the joint is transformed into linear sliding between the two sets of electrodes by a rack-pinion transmission system. (Left) The major subcomponents of the joint. Illustrations both with and without the outer casing are shown. (Right) Rotational motion of the pinion causes the two sets of brakes to linearly slide relative to each other. Conversely, blocking relative motion between the two sets of electrodes will prevent rotation of the joint.}
  \label{fig:joint_design}
\end{figure*}

\subsection{Scalable Electrostatic Force}

The quadratic relationship between braking force and voltage in eq.~(\ref{eq:F2V}) encourages the use of high voltages in order to maximize braking capability. In practice, the applied voltage is limited either by the voltage tolerance of the control electronics, or the dielectric breakdown of the insulator. Generally, the voltage tolerance of the electronics increases with their size, and the insulator’s breakdown voltage is proportional to its thickness (which itself affects the maximum braking force). The amplitude of the applied voltage has practical limits, but there are other avenues of brake design that can further increase braking capability.

First, the proportional relationship between dielectric constant and braking force motivates the use of high relative permittivity insulators ~\citep{hinchet2020high,diller2016lightweight}. These materials are expensive and difficult to source, particularly in comparison to lower permittivity insulators such as polyimide and polyethyleneterephthalate (PET) films. Second, the braking force’s inverse quadratic relationship with plate distance encourages the use of thinner dielectrics, but practical limitations include material availability, fabrication capability, mechanical robustness, and the resulting decrease in breakdown voltage threshold. Third, increasing the area of the electrodes (and their area of overlap) will yield greater braking capability, but will generally make the brake less compact. 

Finally, \rev{non-ideal behavior of the electrodes} can hinder braking strength. In particular, the effective area of overlap between the brake’s electrodes is heavily affected by the degree to which they conform to each other during engagement. Conformance of thicker electrodes is limited by their lack of compliance. On the other hand, thinner electrodes with greater compliance are more susceptible to both unzipping and buckling \rev{(Fig. \ref{fig:non_ideal})}. Unzipping occurs when out-of-plane forces peel the electrodes away from each other at the edges of contact, and buckling occurs when in-plane forces cause one or more electrodes to cave in on themselves~\citep{karagozler2007electrostatic,diller2016lightweight}.  Buckling can be prevented by applying tension to the electrodes when mounted to the brake substrate, and peeling can be mitigated by ensuring that each electrode is limited to motion along its plane. While it is possible to produce a braking effect between electrodes with relative rotational motion~\citep{aukes2014design,johnsen1923physical}, our own experience with alternative joint prototypes suggests that linear sliding best avoids electrode conformance issues. We found that limiting motion between the electrodes to be along a single linear axis eliminates the out-of-plane forces that cause unzipping, and buckling can be prevented by tensioning the electrodes along the direction of motion.

In formulating the design of our brake, we aim to maximize braking strength while minimizing size and cost. While it has a low dielectric constant relative to other materials, we use PET films because they are thin, mechanically robust, and widely available. To minimize the area of our brake's electrodes, we stack multiple brake electrodes on top of each other according to the required maximum braking force. Since electrostatic brakes are constructed from thin materials, stacking them does not significantly increase volume nor weight. Finally, our brake design maximizes effective area of overlap by optimizing electrode conformance. As previously discussed, we achieve this by limiting motion of the electrodes to be along a single linear axis. The design of our brake and corresponding joint is further detailed in Section \ref{sec:joint_design}.

\section{Electrostatic Braking for Robot Joints}
\label{sec:joint_design}

Designing a robotic joint that incorporates an electrostatic brake presents a number of challenges. On top of the requirements for significant braking force generation, it is crucial that a brake-aided joint has consistent braking capabilities throughout its joint range. Achieving such consistency requires the spatial relationship between the conductors (particularly area of overlap, conformance, and separation distance) to remain constant throughout all possible positions of the joint. The design should also be amenable to stacking brakes so that the braking force can be scaled as necessary. Finally, the use of high voltages (on the order of 1 kV) to actuate the brake can cause time-dependent, undesirable effects that are not captured by the parallel plate capacitor model. 

In the following subsections, we describe the design and operation of our brake-aided joint. Our electrostatic brake equipped joint converts rotational motion of the joint into linear sliding motion between the electrodes in order to maintain the optimal and consistent electrode conformance that is assumed by our electrostatic braking model (Sec. \ref{sec:cap_model}). We detail the materials and methods used to construct the joint. Finally, we discuss how injection of charge into the dielectric over time is detrimental to brake strength, disengagement, and re-engagement, and we describe our implementation of control circuitry that \rev{prevents} charge injection.

\begin{figure}[t]
\centering
\includegraphics[width=0.5\textwidth]{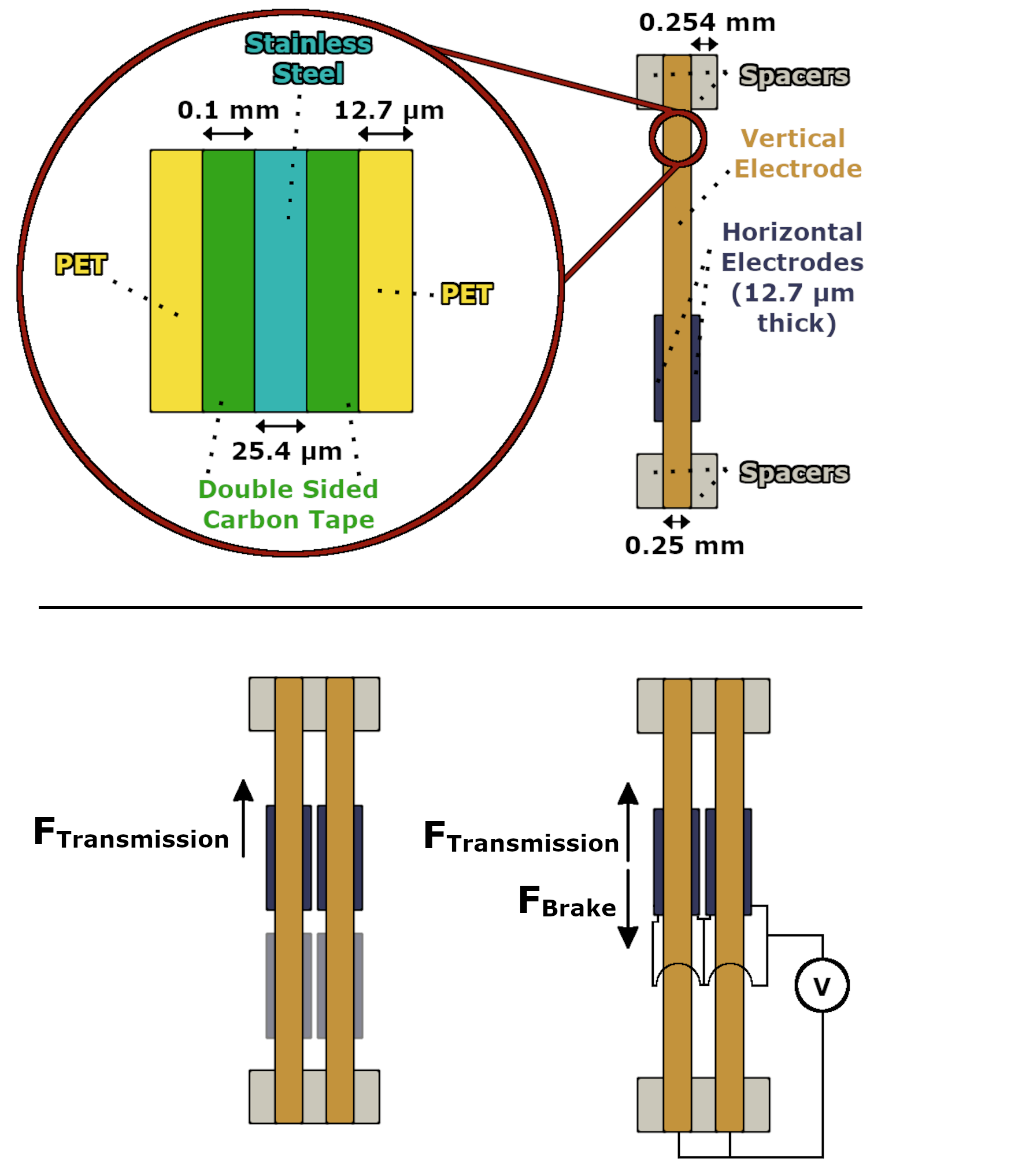}
\caption{ Construction of electrostatic brakes from thin, lightweight materials facilitates stacking in order to increase braking capability. (Top) Side view of a single stack of electrodes consisting of a vertical electrode sandwiched between two horizontal electrodes. (Bottom) Two stacks of electrodes. Spacers separate adjacent stacks of electrodes.  (Bottom Left) When the joint rotates, a linear force causes the horizontal electrodes to slide along the vertical electrodes. (Bottom Right) Application of voltage induces an equal and opposite frictional force that prevents sliding between the electrodes and therefore rotation of the joint.
}
\label{fig:brake_stackup}
\end{figure}

\subsection{Brake-Aided Joint Design}

We first design a modular robot joint that demonstrates the potential of electrostatic brake-aided systems (Fig. \ref{fig:joint_design}). Our joint uses a rack and pinion transmission system to convert the rotational motion of the joint into linear sliding motion between the brake’s electrodes. In our joint design, a set of electrodes mounted to the rack (i.e., the horizontal electrodes) are interleaved with another set of electrodes mounted to the joint’s outer casing (i.e., the vertical electrodes). When the brake is not engaged, rotation of the pinion pushes the rack along stainless-steel rails that have been mounted to the outer casing of the joint. Here, the two sets of electrodes slide along each other with negligible frictional force. Once the brake is engaged, the resulting electrostatic force between the two sets of electrodes prevents them from sliding relative to each other. The rack-mounted set of electrodes thereby prohibits the rack from sliding along the rails, which in turn blocks rotation of the pinion.

We define a stack to be a vertical electrode sandwiched between two horizontal electrodes (Fig. \ref{fig:brake_stackup}). The vertical electrode is composed of a stainless-steel strip that uses double-sided carbon tape to attach a dielectric sheet to each of its faces. Carbon tape is an adhesive solution that does not increase the effective thickness of the dielectric due to its conductivity~\citep{hinchet2020high,hinchet2018dextres}. Each of the horizontal electrodes is simply a strip of stainless steel. When stacking electrodes, we use \rev{stainless steel} spacers to control the spatial offset between adjacent stacks. Stacking electrodes is an effective method for scaling braking strength without adding significant volume, mass, and cost (Table \ref{tab:joints}).

\begin{table}[t]
\centering
\caption{The volume, mass, and cost of the proposed electrostatic brake equipped joints. Row one corresponds to the prototype joint used when measuring holding torque capability. Row two corresponds to the updated joint used in the ten degree-of-freedom robot. Metrics for these two rows do not include the brake (electrodes and spacers) because the brake’s volume, mass, and cost are dependent on the degree of stacking. Instead, the volume, mass, and cost of a single stack of electrodes and spacers is reported in the third row. }
\label{tab:results}
\resizebox{\columnwidth}{!}{%
\begin{tabular}{lccc}\toprule

                  & Volume ($cm^3$) & Mass ($g$) & Cost (USD)\\
\midrule
Prototype Joint   &  20.0           & 20.2       & \$15.42 \\
Updated Joint     & 25.4            & 24.9       & \$16.18  \\
Brake (per stack) & 0.30            & 2.4        & \$0.61 \\
\bottomrule

\end{tabular}
}
\label{tab:joints}
\end{table}

\begin{figure*}
     \centering
     \begin{subfigure}[b]{0.32\textwidth}
         \centering
         \includegraphics[width=\linewidth]{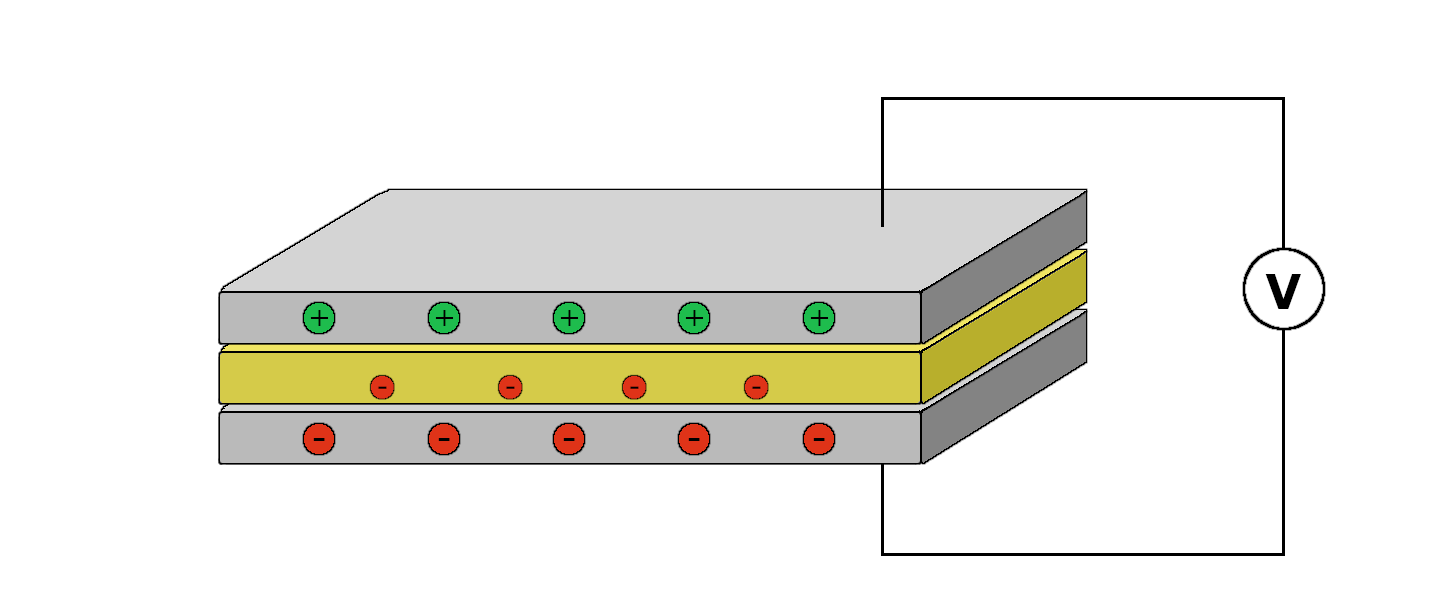}
     \end{subfigure}
     \hfill
     \begin{subfigure}[b]{0.25\textwidth}
         \centering
         \includegraphics[width=\textwidth]{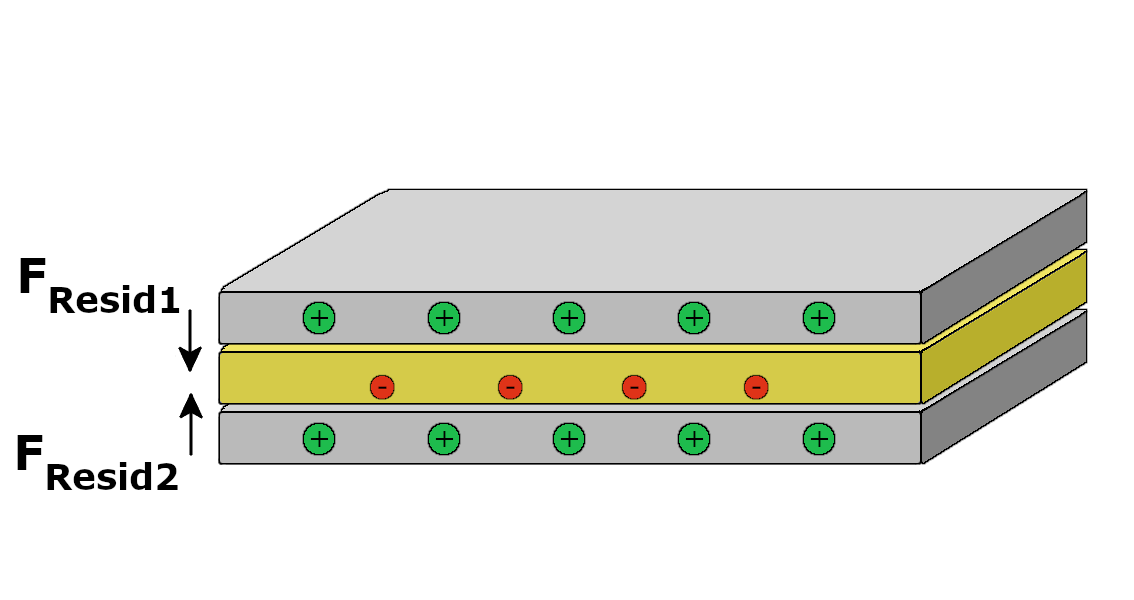}
     \end{subfigure}
     \hfill
     \begin{subfigure}[b]{0.32\textwidth}
         \centering
         \includegraphics[width=\textwidth]{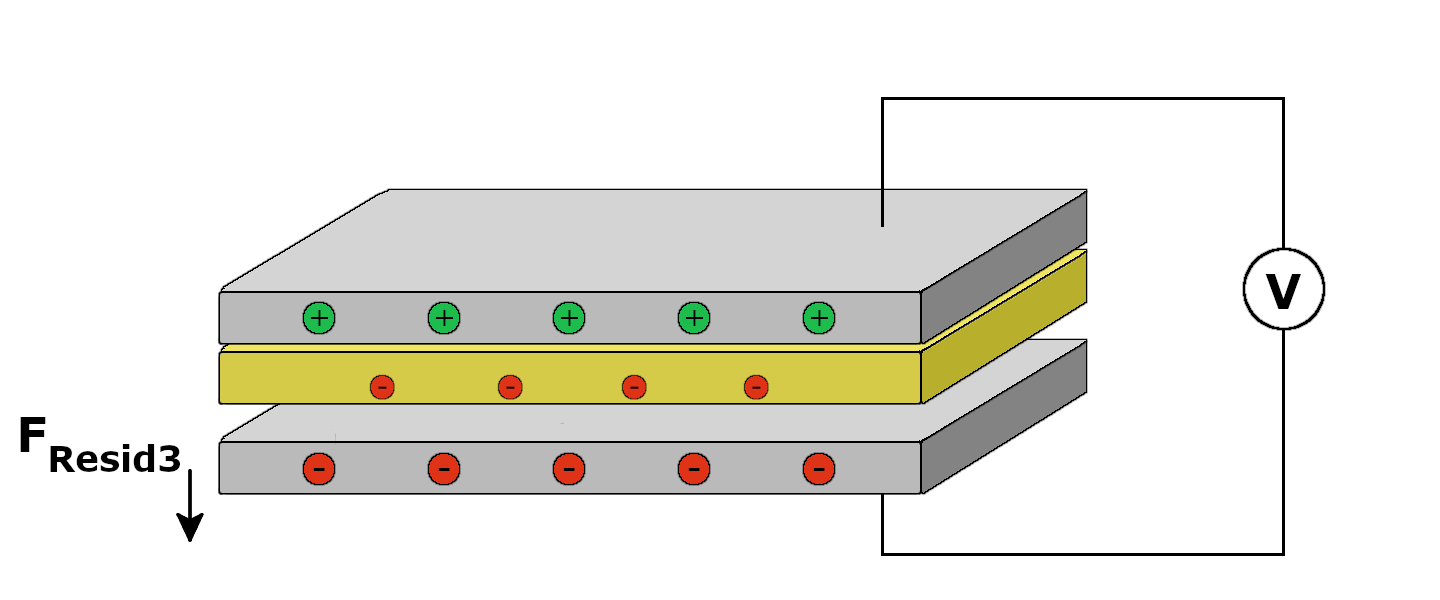}
     \end{subfigure}
        \caption{Charge injection is detrimental to the operation of an electrostatic brake. (Left) Application of high voltage for a sustained period across the brake’s conductors causes electric charges to be injected into the brake’s dielectric. (Middle) Injected charge remains trapped in the dielectric even once the voltage has been removed, preventing the brake from disengaging. (Right) Charges trapped in the dielectric also hinder re-engagement of the brake by generating a repulsive force. }
        \label{fig:charge_injection}
\end{figure*}

\subsection{Joint Construction\label{sec:joint_construction}}

We construct a vertical electrode by first cutting out a 76.2 mm by 8 mm strip of 25.4 µm thick stainless-steel foil with scissors (Fig. \ref{fig:brake_stackup}). Both faces of the same end of this strip are then each covered by a 35 mm by 8 mm strip of double-sided conductive carbon tape. Films of 12.7 µm thick PET dielectric are then laid onto both faces of the vertical electrode such that they cover the carbon tape. Once adhered to the vertical electrodes, a hand-held rotary cutter is used to trim the PET films such that they each have a width of 12 mm. Any film that protrudes beyond the end of the vertical electrode should also be removed using the rotary cutter. 

Horizontal electrodes consist of a 76.2 mm by 10 mm strip of 12.7 µm thick stainless-steel foil. Four square holes that align with mounting holes on the gear rack are cut out of the strip using an \rev{X-Acto} knife. We also fold the horizontal edges of the electrode over itself to ensure that the edges are smooth. This smoothness prevents the edge of the horizontal electrode from cutting into the dielectric layer of the vertical electrode during engagement.

The outer casing, pinion, and rack are 3D printed from the ABS filament of a Stratasys F120 3D printer. \rev{While 3D printed parts have sufficiently tight tolerances and are conducive to rapid prototyping and low cost, future designs could use machined metal parts to achieve better gear meshing.} Both ends of the pinion are mounted to the joint’s outer casing via two ball bearings. \rev{A stainless steel washer is pushed into each of the four holes (two on top, two on bottom) of the plastic rack using a soldering iron.} These washers have a nominal inner diameter of 0.094 inches. They facilitate linear sliding of the rack against two 0.09375 inch diameter dowel pins \rev{(i.e. linear rails)} inserted into the outer casing of the joint \rev{(Fig. \ref{fig:joint_design})}. \rev{The washers' thickness of only 0.02 inches results in small contact area between the washers and linear rails, allowing the rack to slide with minimal friction absent any applied braking forces.}

Stacks of brakes are mounted to the back of the outer casing. Thin and rigid electrical wire is temporarily inserted into the four mounting holes of the rack and the four mounting holes at the top and bottom of the outer casing. These wires help hold the spacers and electrodes in place as they are being stacked on top of each other. A single stack of brakes is formed by first placing a pair of 11.5 mm by 4 mm strips of 0.254 mm thick stainless steel through the wires protruding from the rack and a pair of 21.7 mm by 4 mm strips of 0.254 mm thick stainless steel through the wires protruding from the top and bottom of the outer casing. These spacers are constructed using a hand-held sheet metal cutter and a 2.5 mm diameter metal hole punch. A horizontal electrode is then laid across the rack, followed by a vertical electrode across the top and bottom of the casing, and finally followed by another horizontal electrode across the rack. If this is the final stack of brakes, then the stack’s final layer consists of another pair of spacers at the top and bottom of the outer casing and another pair of rack mounted spacers. If additional stacks of electrodes will eventually be added, then the stack's final layer consists of only a pair of rack mounted spacers. This process can then be repeated in order to add more stacks of brakes. Once any additional stacks of brakes are placed on top, 3D printed fasteners are slipped onto the protruding wires. The wires are then removed and replaced with M2 screws that are twisted into hex nuts embedded in the rack and outer casing in order to hold the stacks of electrodes in place.  

Note that one layer of spacers separates adjacent vertical electrodes in order to ensure that there is space in between them for the horizontal electrodes to slide \rev{(Fig. \ref{fig:brake_stackup})}. The thickness of horizontal electrodes is negligible in comparison to that of vertical electrodes. Therefore, by choosing stainless steel spacers with a thickness as close as possible to that of a vertical electrode, two layers of spacers between adjacent stacks of horizontal electrodes will result in an offset that is approximately equal to the offset between adjacent vertical electrodes. The closer these offsets are, the flatter the contact between the vertical electrode and horizontal electrodes within a given stack, which is conducive to electrode conformance and effective area of overlap.


\subsection{Charge Injection}
\label{sec:charge_injection}
The application of a high \rev{DC} voltage to the brake during engagement significantly increases the amount of time required to subsequently disengage the brake~\citep{aukes2014design}. As shown in Fig. \ref{fig:charge_injection}, when a high \rev{DC} voltage is applied across the brake’s plates, electrical charges are injected into the brake’s dielectric and become trapped inside~\citep{montanari2000electrical,dissado1997role}. These stored charges induce a residual attractive force between the plates that prevents brake disengagement after the removal of voltage and hinders brake re-engagement once voltage is again applied. The magnitude and persistence of this residual force increases with the duration of the preceding applied voltage~\citep{aukes2014design}. This observation suggests that the applied voltage signal during engagement take the form of a bipolar square wave. This waveform maintains the full voltage magnitude across the plates, but also periodically inverts the voltage polarity such that the effect of charge injection is made negligible~\citep{hinchet2020high,hinchet2018dextres}.


\begin{figure}[t]
\centering
\includegraphics[width=0.5\textwidth]{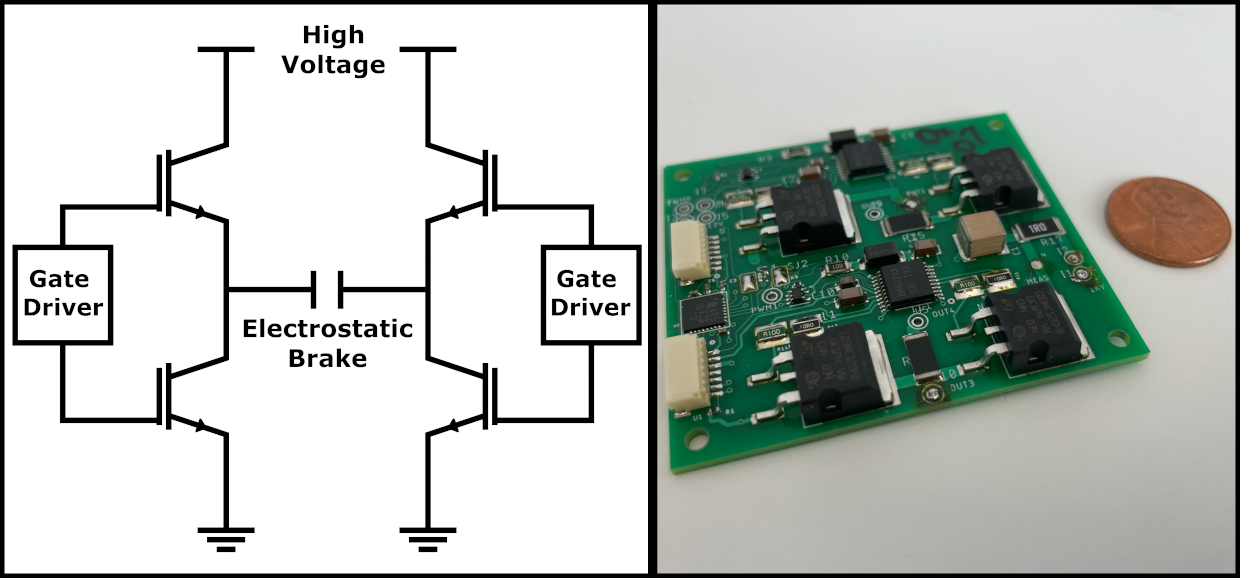}
\caption{ To mitigate charge injection, the polarity of the applied voltage must be periodically inverted during brake engagement. (Left) Schematic of the high voltage H-bridge circuit used to control the brake. (Right) The printed circuit board that implements the H-bridge circuit.
}
\label{fig:brake_board}
\end{figure}

\subsection{Brake Driver Circuitry}

In order to actuate the brakes, we designed a printed circuit board for generating high voltage bipolar square waves (Fig. \ref{fig:brake_board}). The circuit implements an H-bridge architecture to enable swapping of the voltage polarity across its two outputs. Our 1200V tolerant H-bridge is composed of four IGBT transistors controlled by two gate driver modules. Given a single joint, all of the joint’s horizontal electrodes are connected to one output of the H-bridge, while all of its vertical electrodes are connected to the other output.

\section{Brake-Aided Joint Performance}
\label{sec:brake_performance}
In this section, we evaluate the performance of our brake-aided joint and demonstrate how it can enhance the dexterity of highly articulated, underactuated systems without significantly increasing weight, volume, or power consumption. Here, we aim to quantify the conformance of the brake's electrodes throughout the range of the joint. Specifically, we measure the maximum braking torque that the brake-aided joint can exert, and compare it to the performance predicted by the ideal-conformance, parallel plate capacitor model. We expect that the observed maximum braking torque will increase linearly with the degree of stacking. We then compare these results to the holding torques of motors commonly used in highly articulated robots to establish that electrostatic brakes can serve as \rev{an alternative to electromechanical motors when a lighter, compact and more power efficient solution is required without the need for full actuation}. 


For the remainder of this section and those that follow, we use an applied voltage of 1000V,  PET insulator with approximate permittivity of 3.35~\citep{mcmahon1959degradation} and thickness 12.7 µm, and a stainless steel-PET coefficient of friction value of 0.71~\citep{mens1991friction}. The only exception is that we used a nominal voltage of 950V when performing the multi-object manipulation task with our ten degree-of-freedom robot and during in-hand manipulation with our two-fingered robot. We lowered the voltage because the output of our high voltage power supply is dependent on the load current. In the most extreme case of all brakes turning on or off at the same time, the output voltage could momentarily fluctuate by up to 100V. Lowering the nominal output voltage ensured that the voltage tolerance of our H-bridge circuit was not exceeded when these fluctuations occurred. 

\begin{figure}[t]
\centering
\includegraphics[width=0.48\textwidth]{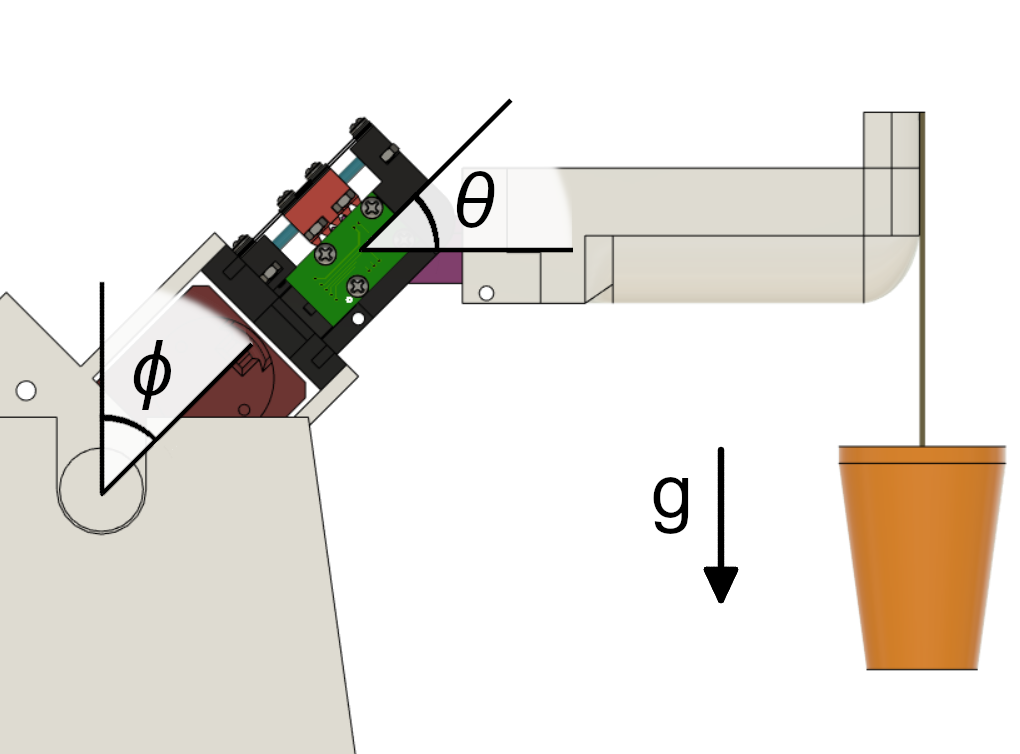}
\caption{ Visualization of the experiment platform that facilitates torque measurements throughout the joint’s range. For a desired test setting of the joint angle $\theta$, the base angle $\phi$ is chosen to orient the lever arm perpendicular to the direction of gravity. Torque applied to the joint is increased by adding known weights to the orange cup attached to the end of the lever arm.
}
\label{fig:torque_measurement}
\end{figure}

\begin{figure*}
  \includegraphics[width=0.95\textwidth]{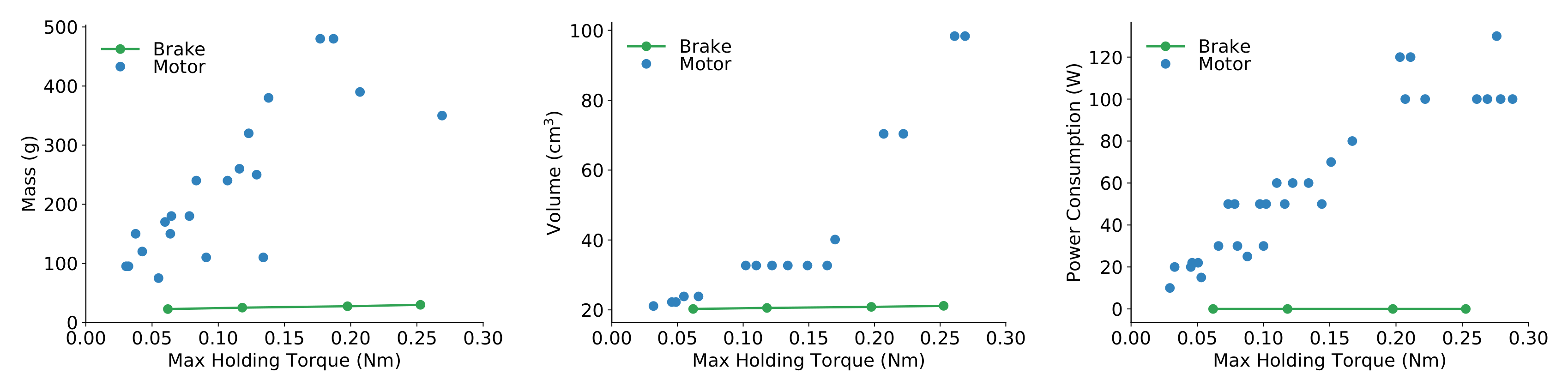}
  \caption{Comparison of the brake equipped joint with motors that have similar holding torques across multiple metrics. \rev{Note that each blue dot corresponds to a particular motor from a catalog of existing Maxon motors. This catalog directly specifies holding torque and mass, whereas volume and power consumption are derived from cylindrical radius/height specifications and nominal voltage and holding torque current specifications respectively. For each individual metric, we generally report motors that are best performing on that metric (i.e. towards the lower-right of the plot).} Our joint is at least four times lighter and one thousand times more power efficient than the compared motors. (Left) Mass comparison. (Middle) Volume comparison. (Right) Power consumption comparison. }
  \label{fig:motor_compare}
\end{figure*}

\begin{figure}[t]
\centering
\includegraphics[width=0.5\textwidth]{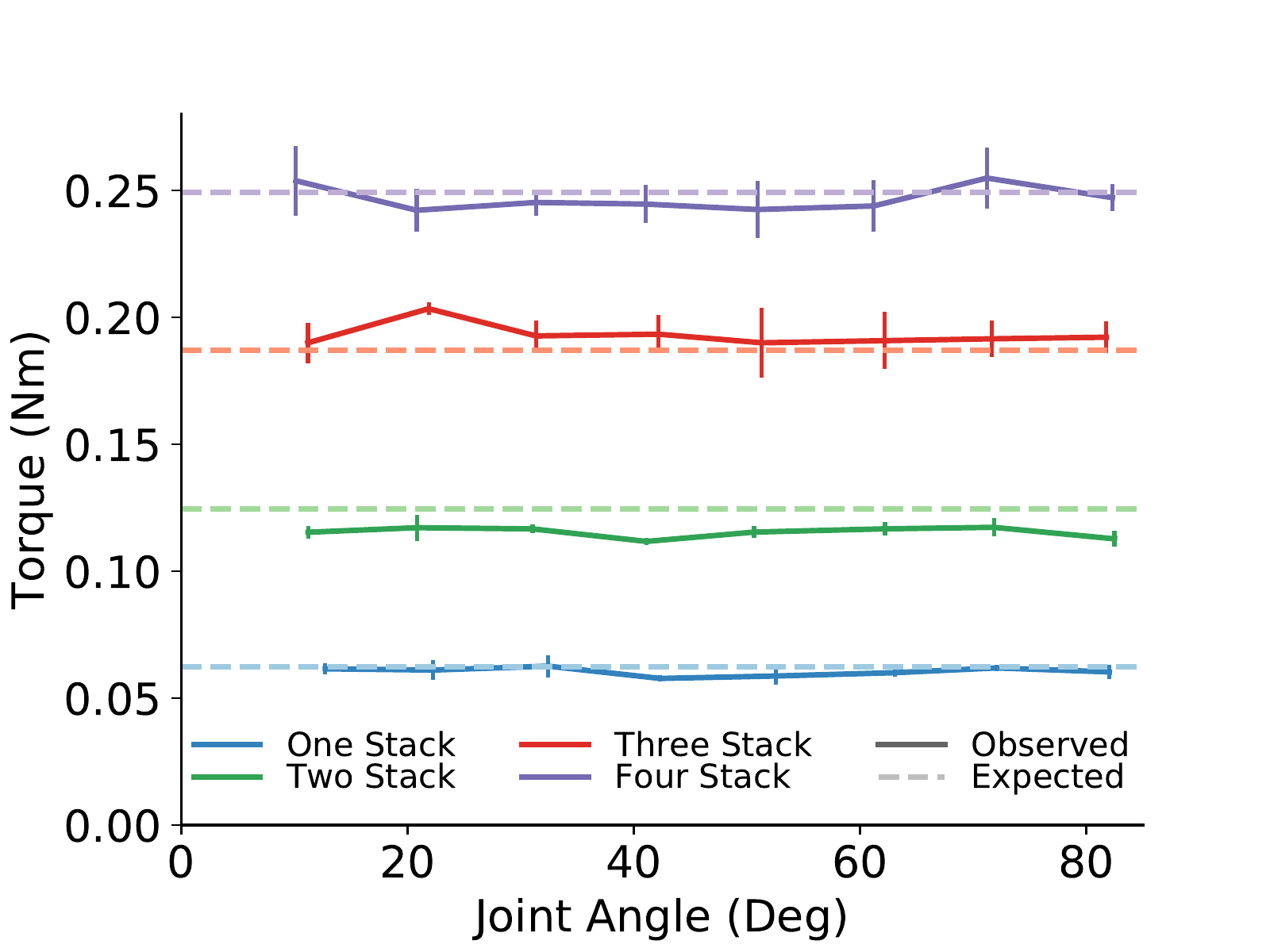}
\caption{ Observed and expected holding torque measurements throughout the joint’s range of motion. Configurations of up to four stacks of electrodes are measured. Measurements are averaged over five trials.
}
\label{fig:few_electrodes}
\end{figure}

\subsection{Electrostatic Brake Stacking}

To verify the consistency and scalability of braking strength, we evaluate our joint’s maximum holding torque throughout its joint range for an increasing number of stacked electrodes (Fig. \ref{fig:torque_measurement}). Measurements consist of applying increasing amounts of torque to the pinion until the engaged brake can no longer prevent relative motion between the two sets of electrodes, causing the joint to rotate. The expected holding torques can be computed using the previous force equations and the rack-pinion equation:

\begin{equation}
    T_{max\:brake} = \frac{1}{2} d_{Pinion} F_{max\:brake}\mbox{,}
\end{equation}
where the pitch diameter of the pinion $d_{Pinion}$ is 12 mm and each vertical electrode - horizontal electrode pair is expected to have an approximate area of overlap of 0.8 cm\textsuperscript{2} throughout the joint's range of motion. 

The applied torque was computed by applying a known weight to a fixed lever arm of total length 0.1 meters. The joint was oriented such that the length of the lever arm was perpendicular to the direction of gravity. The weight was increased in increments of approximately 5 grams by adding small metal pellets to the cup attached to the end of the lever arm. Our joint is equipped with an AEAT-8800 magnetic encoder, which we used to measure the joint angle throughout the experiment. The brake was engaged using a 15 Hz bipolar square wave with an amplitude of 1000 V. \rev{Note that it is necessary to use an AC signal to prevent charge injection, as discussed in Section \ref{sec:charge_injection}}. While this is a high voltage, we used a power supply limited to a maximum current output of 1 mA as a safety measure. We measured the maximum holding torque at intervals of ten degrees for joints with up to four stacks of electrodes. For each configuration of joint angle and stacks of electrodes, we performed five measurements of the holding torque. 

As expected, the brake's holding torque increases proportionally to the number of electrode stacks (Fig. \ref{fig:few_electrodes}). Small differences between the observed and expected torque were likely caused by slight deviations in dielectric thickness, electrode fabrication, and applied voltage. Furthermore, the joint’s braking capability is generally independent of joint angle, which suggests that electrode conformance is consistent throughout the joint’s range of motion. We expect that perpetually increasing the degree of stacking will continue to yield linear increases in braking torque as long as the spacing between adjacent stacks can be kept constant.

For holding torques within operating ranges of robotics applications, electrostatic brakes have significantly less weight, size, and power consumption than motors, which serve as the primary mode of actuation in robotics (Fig. \ref{fig:motor_compare} and Table \ref{tab:joints}). Here, we compare our electrostatic brake equipped joint to Maxon motors because they are frequently used (either directly or as a subcomponent of a servo motor system) in high degree-of-freedom robots~\citep{liu2008multisensory,wright2012design,tuffield2003shadow,ha2011development,buckingham2007snake,schulz2011first,liljeback2017eelume,karakasiliotis2013legged}. We observe that our brake equipped joint is at least four times lighter and one thousand times more power efficient (approximately $0.01 W$ power consumed per stack) than the compared motors. Furthermore, our joint exerts the same holding torques as the larger volume motors while remaining as compact as the smaller motors. Yet we also recognize that motors have capabilities that brakes do not; brakes can only resist force/torque, while motors can independently apply force/torque. However, if the use of motors in a particular application violates design constraints, one or more of them can be replaced by our brake in order to achieve the same amount of holding torque while significantly reducing weight, volume, and power consumption.

Our measurements for the mass, volume, and cost of the brake equipped joint include that of the outer casing, rack, pinion, and brakes, as well as that of the materials for coupling them together (ball bearings, dowel pins, spacers, hex nuts, and screws). By measuring the voltage drop across a current sense resistor in series with the output of the high voltage power supply and the H-bridge circuit, we computed the current consumption of the electrostatic brake with Ohm’s law. Given the output voltage of the power supply, we used these current measurements to compute the brake's power consumption. The holding torque, mass, volume, cost, and power consumption of the motors are either directly given or derived from the Maxon catalog. For each individual metric, the best performing motors over the observed range of braking torques were reported.

\section{Braking for Highly Articulated Mechanisms}
\label{sec:ten_degree_robot}
Any series of two or more robotic joints can benefit from the reduced volume, weight, and power consumption associated with exchanging motors for electrostatic brakes. These benefits become particularly evident for mechanisms that leverage many degrees of freedom to achieve their task, such as wrapping around objects. For example, reducing the power consumption of a deep-sea robotic tentacle will improve its battery life, or decreasing the weight of a snake robot will allow it to climb more effectively. To demonstrate the value of electrostatic braking for such mechanisms, we developed a robot composed of a serial chain of ten brake equipped joints, driven by only a single servo motor. Despite only having a single motor, these brakes facilitate the execution of highly dexterous maneuvers.

\begin{figure}[t]
\centering
\includegraphics[width=0.5\textwidth]{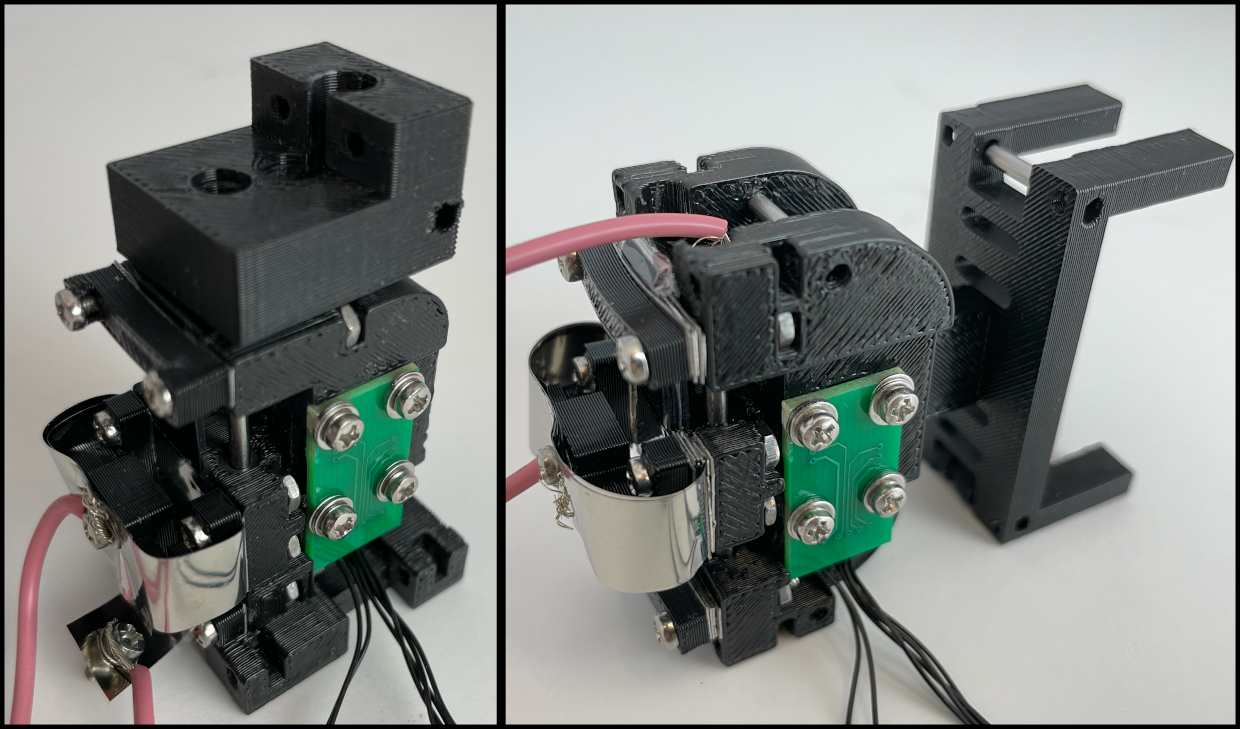}
\caption{The joint design can be adapted according to the desired robot kinematics. (Left) Original joint design used to evaluate braking strength and as the distal joints of the robot hand. (Right) A joint \rev{modified to have} symmetric joint limits \rev{is} used in the ten degree-of-freedom robot and the proximal joints of the robot hand.
}
\label{fig:wide_joint}
\end{figure}

\begin{figure}
     \centering
     \begin{subfigure}[b]{0.3\columnwidth}
         \centering
         \includegraphics[width=\linewidth]{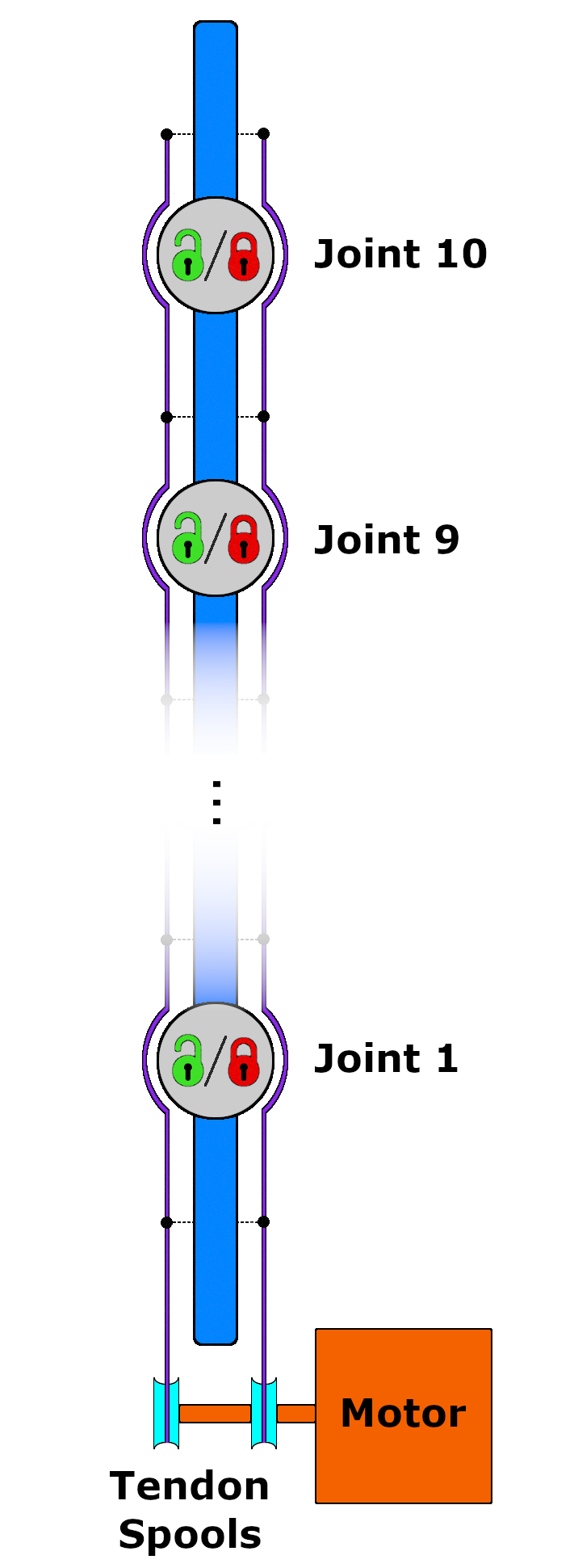}
     \end{subfigure}
     \hfill
     \begin{subfigure}[b]{0.6\columnwidth}
         \centering
         \includegraphics[width=\textwidth]{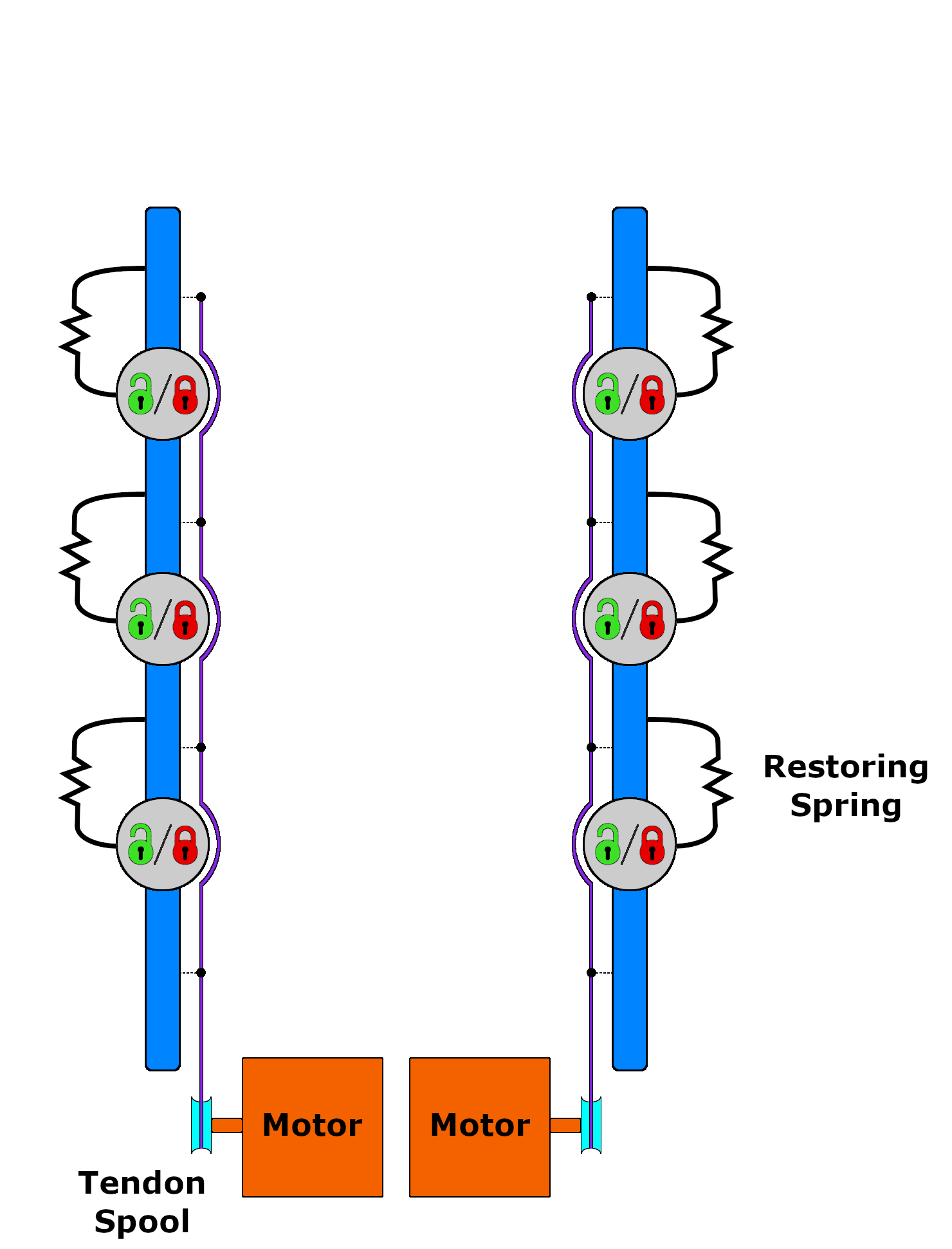}
     \end{subfigure}
        \caption{\rev{The mechatronic structures of the ten degree-of-freedom robot (Section \ref{sec:ten_degree_robot}) and two-fingered hand (Section \ref{sec:hand_robot}) The braking of each individual joint can be controlled independent of all other joints. (Left) A single motor actuates a pair of tendons coupled to all ten degrees-of-freedom. As one tendon is reeled in by the motor, the other tendon is released. Reeling in the left tendon causes all unbraked joints to experience a counter-clockwise torque, while reeling in the right tendon causes them to experience clockwise torque. (Right) Each of the hand's fingers consist of three joints that are all coupled by a tendon actuated by a single motor. Here, the motor reels in the tendon to flex the fingers inwards, and releases the tendon to allow  springs embedded in the joints to extend the fingers towards their home position.}}
        \label{fig:robot_kinematics}
\end{figure}

\begin{figure*}
     \centering
     \begin{subfigure}[b]{0.244\textwidth}
         \centering
         \includegraphics[width=\linewidth]{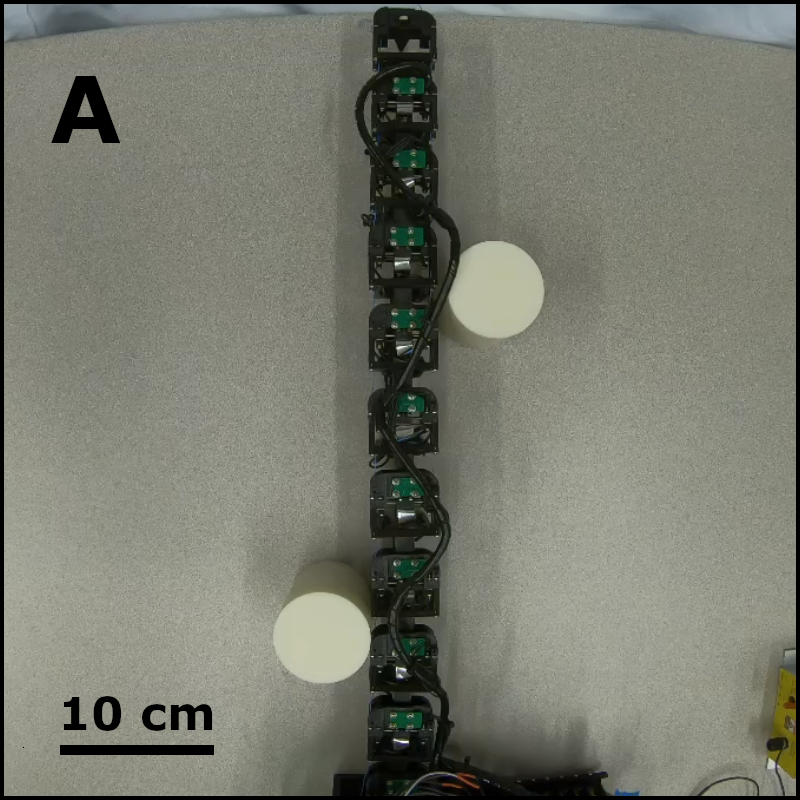}
     \end{subfigure}
     \hfill
     \begin{subfigure}[b]{0.244\textwidth}
         \centering
         \includegraphics[width=\textwidth]{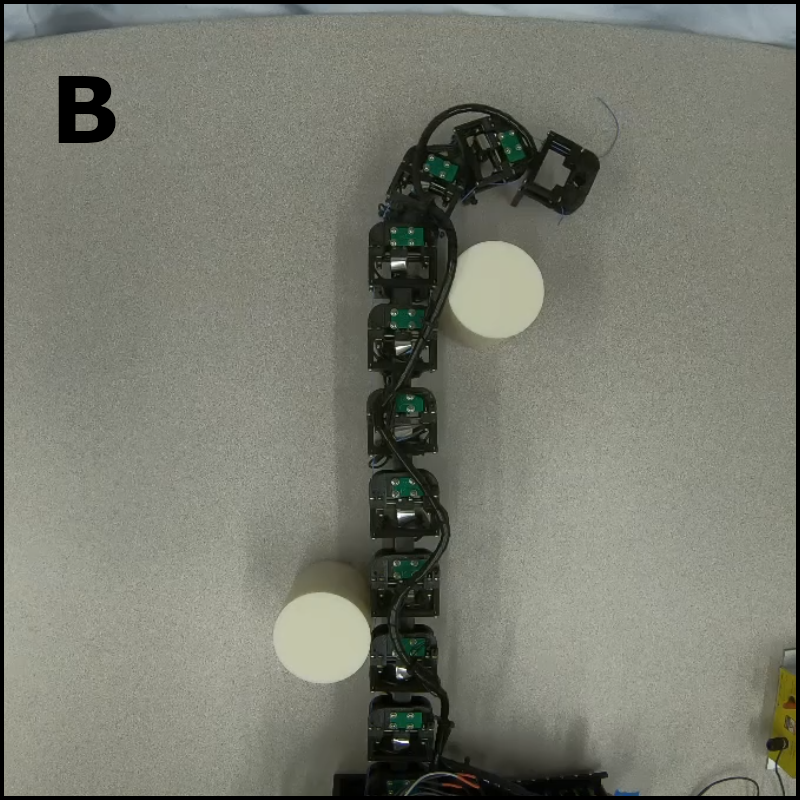}
     \end{subfigure}
     \hfill
     \begin{subfigure}[b]{0.244\textwidth}
         \centering
         \includegraphics[width=\textwidth]{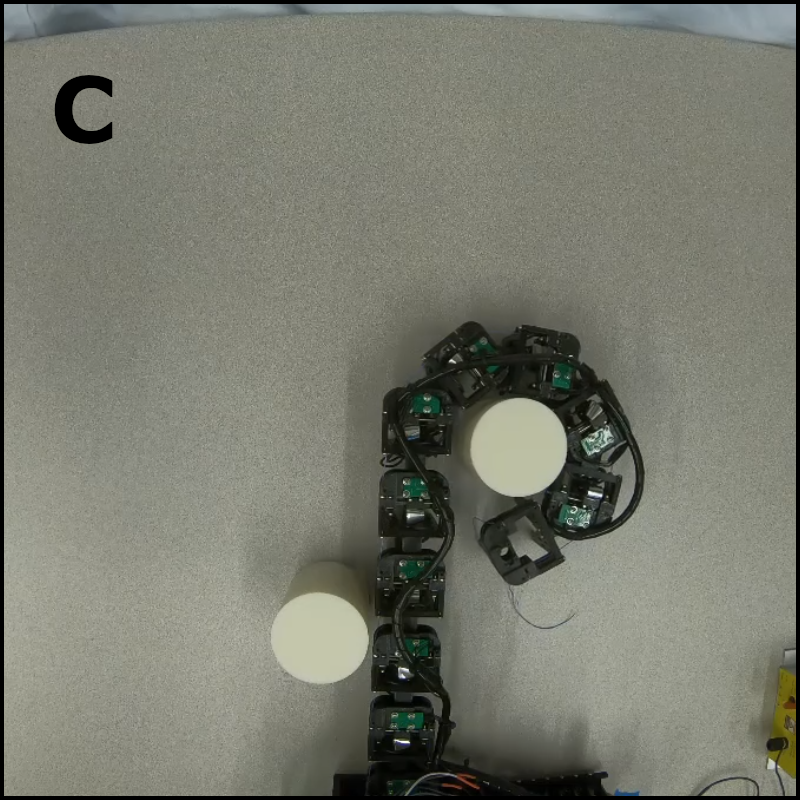}
     \end{subfigure}
     \hfill
     \begin{subfigure}[b]{0.244\textwidth}
         \centering
         \includegraphics[width=\textwidth]{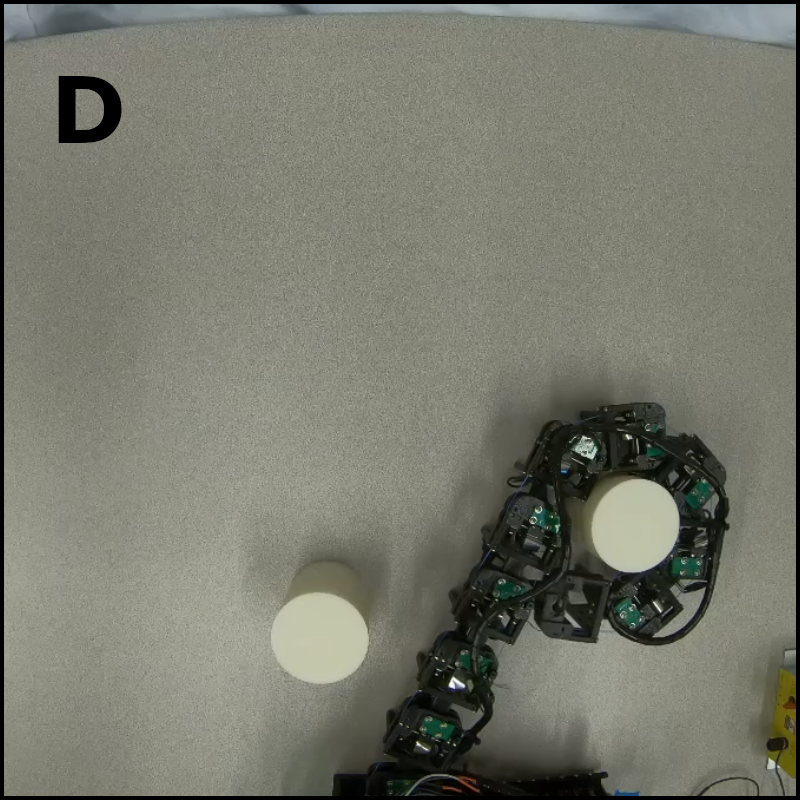}
     \end{subfigure} 
     
     \smallskip
     
     \begin{subfigure}[b]{0.244\textwidth}
         \centering
         \includegraphics[width=\linewidth]{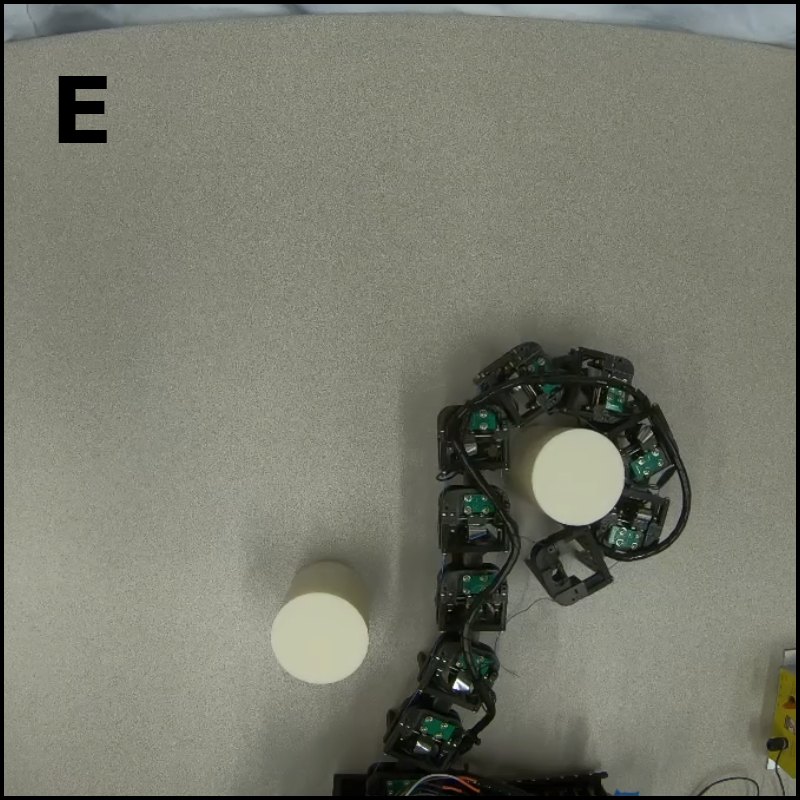}
     \end{subfigure}
     \hfill
     \begin{subfigure}[b]{0.244\textwidth}
         \centering
         \includegraphics[width=\textwidth]{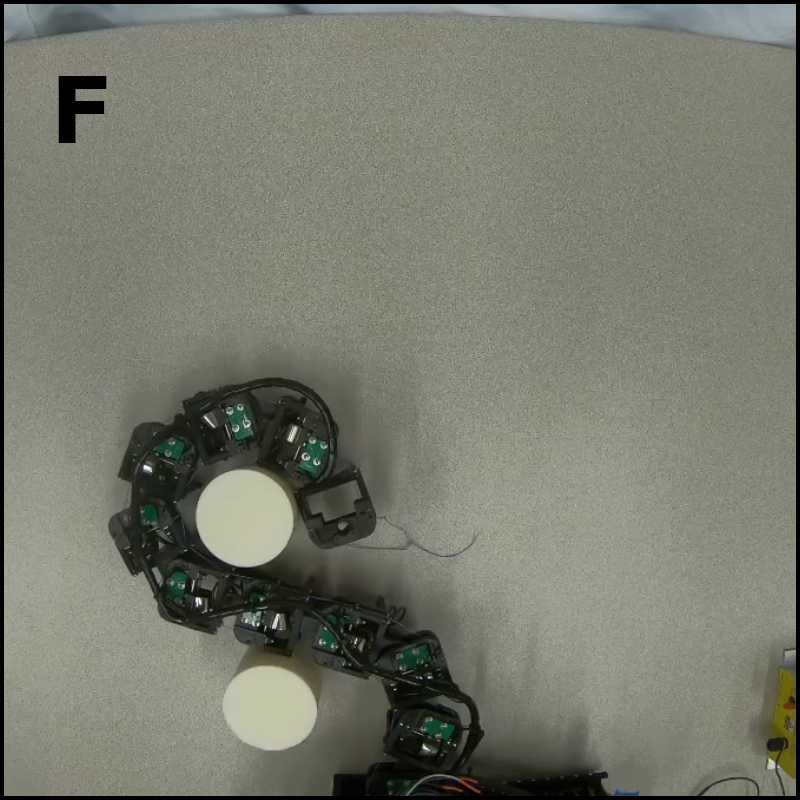}
     \end{subfigure}
     \hfill
     \begin{subfigure}[b]{0.244\textwidth}
         \centering
         \includegraphics[width=\textwidth]{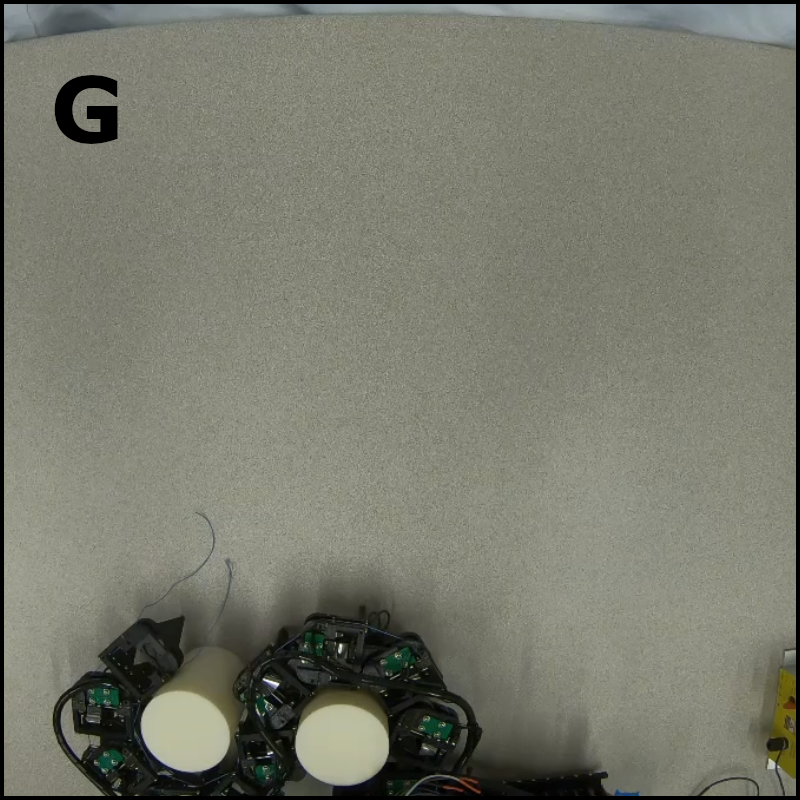}
     \end{subfigure}
     \hfill
     \begin{subfigure}[b]{0.244\textwidth}
         \centering
         \includegraphics[width=\textwidth]{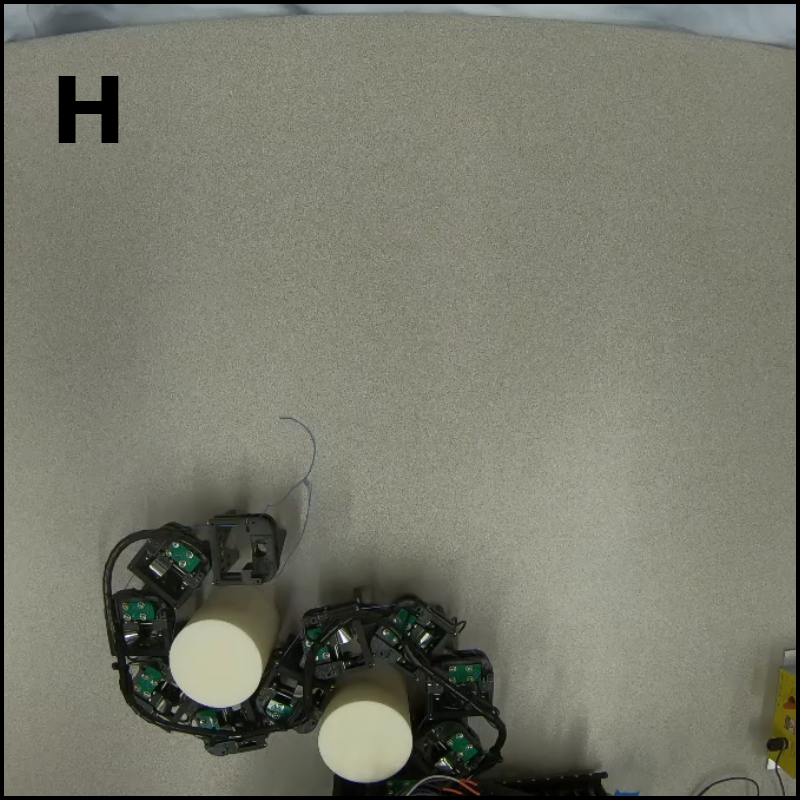}
     \end{subfigure}       
        \caption{The ten degree-of-freedom, single servo motor robot wraps around two objects and then translates them along the tabletop. (A) The initial configuration of the robot and objects. (B) The robot begins wrapping around the upper cylinder. (C) The robot completes a cage around the top cylinder. (D) The robot re-positions in order to avoid a self-collision during its upcoming cage of the lower cylinder. (E) The robot translates the upper cylinder back towards the center of the table. (F) The robot maintains its cage around the upper cylinder as it begins to wrap around the lower cylinder. (G) The robot forms a cage around the lower cylinder. (H) The robot manipulates both objects by moving its most proximal joint while locking all other joints in order to maintain both cages. }
        \label{fig:demo}
\end{figure*}

\subsection{Robot Description}
\label{sec:chain_desc}
We modified the outer shell of each joint to provide routing for the robot’s single pair of antagonistically driven tendons, prevent inadvertent contact between the brake electrodes and other parts of the robot and its environment, and increase the overall joint range to have symmetric limits of –60° and 60° (Fig. \ref{fig:wide_joint} and Table \ref{tab:joints}). The robot’s brake design did not change, and the description of joint construction in Subsection \ref{sec:joint_construction} still applies here. Each joint was equipped with two stacks of brake electrodes. 

Our ten degree-of-freedom robot uses a single Dynamixel XM430 servo motor to exert tension on a single pair of tendons that are routed through the robot’s joints \rev{(Fig. \ref{fig:robot_kinematics})}. Each tendon is a segment of 0.45 mm diameter braided fishing line that can withstand up to 65 pounds of tension. One tendon is routed through the left side of the robot, while the other is routed through the right side. The distal end of each tendon is secured to the final link of the robot by wrapping it around a screw that has been twisted into a screw insert. The proximal end of each tendon is wrapped around its own dedicated 3D-printed tendon spool. The single servo motor controls the motion of both spools; as one spool is reeled in by the motor, the movement of the robot unreels the other spool as necessary.


\rev{We assume that the robot is given a sequence of joint configurations that corresponds to the execution of some desired task. For this experiment, the sequence of joint configurations was manually specified for a known set of object locations; future work could employ object detection algorithms to estimate the pose of object(s) to manipulate and then use motion planning to find an appropriate joint configuration sequence.} In order to reach a desired joint configuration, our ten degree-of-freedom robot controls the state of its ten brakes (on or off) and the velocity of the single motor. For this experiment, the magnitude of the commanded velocity of the motor is always 0.2 rad/sec. To decide the sign of the velocity, each joint votes for the sign corresponding to the motor velocity required to reach its desired joint value. The joints that are in the minority turn on their brakes, and the motor’s commanded velocity is set according to the votes of the majority. As each joint in the majority reaches its desired joint value, it turns on its brake. Once all of the joints in the majority have reached their desired values (and therefore have turned their brakes on), the joints in the minority turn off their brakes and the motor’s commanded velocity is set according to the wishes of the brakes in the minority. As each joint in the minority reaches its desired joint value, it turns on its brake. Once the final joint in the minority reaches its desired value, the overall joint configuration has been achieved. \rev{This process is then repeated for any remaining joint configurations in the sequence.}

\begin{figure*}
  \includegraphics[width=\textwidth]{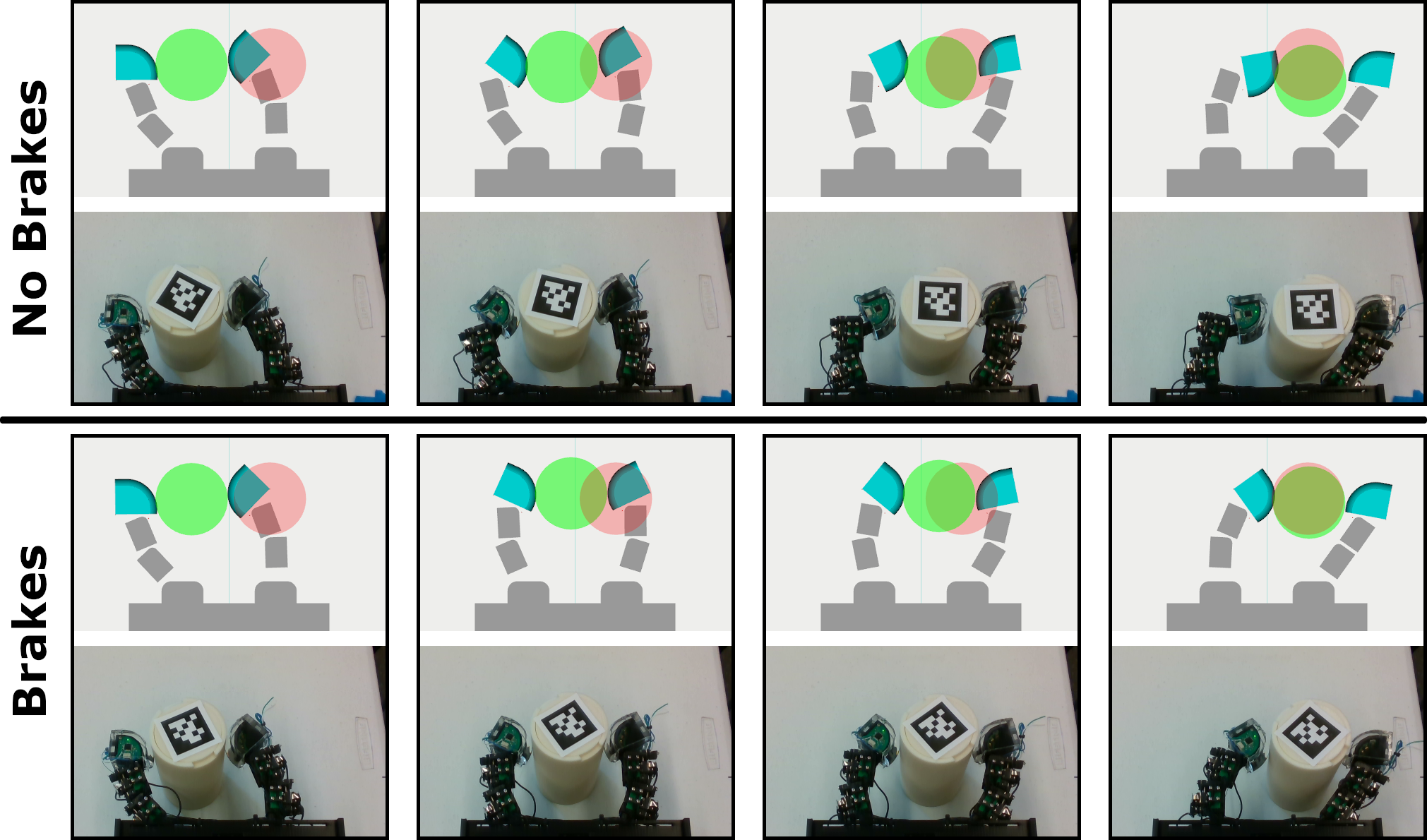}
  \caption{Representative in-hand manipulation trials both with and without the use of electrostatic brakes. For each pair of images, the upper image shows a visualization of the robot-object system, and the lower image shows the real object and robot. The green cylinder  represents the pose of the object, and the red cylinder represents the desired goal pose of the object. The left-most snapshots show the initial pose for both trials, right most snapshots show the final pose of the object, and intermediate snapshots show the progression of the manipulations. Top: A manipulation trial in which the brakes are not used. Bottom: A brake-aided manipulation trial. }
  \label{fig:hand_traj}
\end{figure*}

\subsection{Manipulation of Multiple Objects}

As an example, we consider the challenging task of simultaneously manipulating two plastic cylinders (Fig. \ref{fig:demo}). The robot uses its distal links to wrap around the top cylinder. Once the cylinder has been caged, the robot curls its more proximal links around the lower cylinder while maintaining the distal links’ cage. After creating a second cage around the lower cylinder, the robot locks all but the single most proximal of its joints. This allows it to maintain both cages as it actuates the joint at its base in order to manipulate both cylinders. 

\section{Brake-Aided In-Hand Manipulation}
\label{sec:hand_robot}
Electrostatic brakes facilitate improved manipulation speed and precision by enabling the control of individual joints in underactuated mechanisms. In this section, we develop a robot hand composed of two motors and six brake-equipped joints. We evaluate the hand's ability to perform a planar in-hand manipulation task in which it must translate an object from one side of its workspace to the other (Fig. \ref{fig:hand_traj}). During this manipulation task, we observe the robot's performance both with and without the use of brakes. In particular, we measure the hand's ability to position the object at a desired pose and the amount of time required to complete the task.

\subsection{Electrostatic Brake Enabled Hand}

Our two-fingered robot hand uses both of the previously described brake-equipped joint versions. Each finger consists of a symmetric joint at the base, as well as an intermediate prototype joint and a distal prototype joint. Each joint is equipped with two stacks of brake electrodes.

A compliant fingertip is coupled to the end of each finger by the distal joint. Each fingertip is constructed by embedding four optical proximity sensors into the rigid skeleton of the fingertip, and then encasing it inside of polydimethylsiloxane (PDMS) ~\citep{patel2016integrated,lancaster2019improved}. Our previous work demonstrated that the distance measurements from the embedded proximity sensors provide sufficient sensory feedback to localize an object throughout in-hand manipulation ~\citep{lancaster2022inhand}. In order to focus on the performance of the brakes, this work instead uses more conventional pose estimation methods (i.e. visual fiducials) to localize the object. 

The robot hand is actuated by tendons that are driven by two Dynamixel XM430 servo motors \rev{(Fig. \ref{fig:robot_kinematics})}. For each finger, a tendon is anchored to the fingertip and then routed through the medial side of the finger's links. The routed end of the tendon is then attached to a motor dedicated to the movement of the corresponding finger. The finger flexes inwards when the motor pulls on the tendon. For extension, we slip two springs over the two dowel pins of each joint such that the ends of the springs make contact with the gear rack and outer casing. These springs are compressed when the finger is flexed by the motor. When the motor releases the tendon, the springs extend and induce extension of the finger by pushing the gear rack along the dowel pins.

\subsection{Model Predictive Control for In-Hand Manipulation}

We use model predictive control (MPC) to generate control actions for in-hand manipulation as previously reported in ~\citep{lancaster2022inhand}. Our model predictive controller explores the space of possible trajectories in order to find actions that move the system closer to the goal state. We discuss the dynamics model it uses to simulate trajectories that begin at the current system state, and describe the cost function used to measure the quality of a simulated trajectory.

Our controller uses a dynamics model to predict how the state of the robot-object system will evolve in response to the robot's actions. We define the system state $s_t$ as the angular positions and velocities of the robot's six joints and the $xy$ position and velocity of the cylindrical object. The robot's actions $a_t$ are defined as the commanded position of the two motors and the states of each of the six brakes (on or off). In general, any of the 64 possible braking combinations can be executed. For the sake of creating a tractible action space for the controller to explore, we limit the braking actions to the 9 braking configurations in which exactly one of the brakes in each finger is off. In order to generate a large number of trajectories, we simulate our robot in Isaac Gym ~\citep{makoviychuk2021isaac}. This GPU based physics simulator serves as a highly parallelizable dynamics model.

We implement a model predictive controller for our robot hand by modifying the model predictive path integral (MPPI) method to be compatible with our hybrid action space ~\citep{williams2017information, zhong2019}. The original MPPI framework generates a large number of action sequences, produces trajectories and corresponding costs by simulating these action sequences, and then outputs a cost-weighted average of those action sequences for execution on the real robot. However, this averaging is not applicable to a hybrid action space for which the discrete variables lack a Euclidean distance measure. \rev{We rectify this by choosing to} require each action sequence to maintain a consistent brake configuration throughout the corresponding simulated trajectory, and then compute cost-weighted averaged action sequences for each possible brake configuration. Initially, of the 9 outputted action sequences, we execute the first action of whichever sequence has the lowest cost. For subsequent time steps, we only choose an action sequence corresponding to a different brake configuration if it has a significantly lower cost (we use a constant threshold percentage $\phi$) than the cost of the action sequence corresponding to the previous brake configuration. Otherwise, we execute the first action of the action sequence corresponding to the previous brake configuration. 

At each time step, our MPPI controller simulates many trajectories over a time horizon $T$. It computes a cost for each trajectory that penalizes the fingers not making contact with the object and object distance from the goal:

\begin{equation}
    J(s_t, \dots, s_{t+T}) = a_1 \cdot \displaystyle \sum_{\tau = t}^{t+T}  \mathcal{I}(s_{\tau}) + a_2 \cdot \lvert x_{g}-x_{t+T} \rvert 
\end{equation}
where $\mathcal{I}(s_{\tau}) $ is an indicator function that returns the number of fingertips \textit{not} in contact with the object, and $x_{g}$ is the desired goal position of the object.

Both visual fiducial localization and our MPPI controller are simultaneously executed on a single desktop PC with an Intel i7 Quad-Core CPU, 64 GB of RAM, and a NVIDIA Titan XP GPU. Initial values for all of the following parameters were chosen based on our intuition and then hand-tuned until reasonable performance was achieved. Regardless of whether brakes are being used or not, our MPPI controller simulates a total of 297 trajectories over a time-horizon $T=10$ at each time step . Its cost function uses parameter values $a_1 = 0.1$, $a_2 = 200$, and the MPPI hyperparameter $\lambda$ is set to 0.1. Switching between braking configurations is thresholded on a value of $\phi = 25\%$. Controls are generated at a rate of 5 Hz.

\subsection{In-Hand Manipulation Evaluation}

We evaluate the performance of our brake-enabled robot hand during in-hand manipulation. The in-hand manipulation task consists of the robot moving an object from one side of its workspace to the other. The initial position of the object is 4.5 cm to the left of the geometric plane that symmetrically bisects the robot  and 4.5 cm above the base of the fingers (Fig. \ref{fig:hand_traj}). The configuration of the fingers at the beginning of the task was chosen so that the robot is able to complete the manipulation without \rev{needing to turn on any of its brakes}. The goal position of the object is the reflection of the initial position across the bisecting plane.  The manipulation object is a 3D printed cylinder of radius 4 cm and height 14 cm.

In order to measure the advantage of brake enabled mechanisms relative to their conventional underactuated counterparts, we quantify the robot's in-hand manipulation dexterity both with and without the use of brakes. Specifically, we measure the execution time required to complete the manipulation and the distance between the object and the goal position at the end of the manipulation. We perform ten trials for each method (no braking and braking). A trial is terminated and considered a success once the horizontal distance between the object and the goal pose is less than 1 mm.

\begin{figure}
     \centering
     \begin{subfigure}[b]{0.45\columnwidth}
         \centering
         \includegraphics[width=\linewidth]{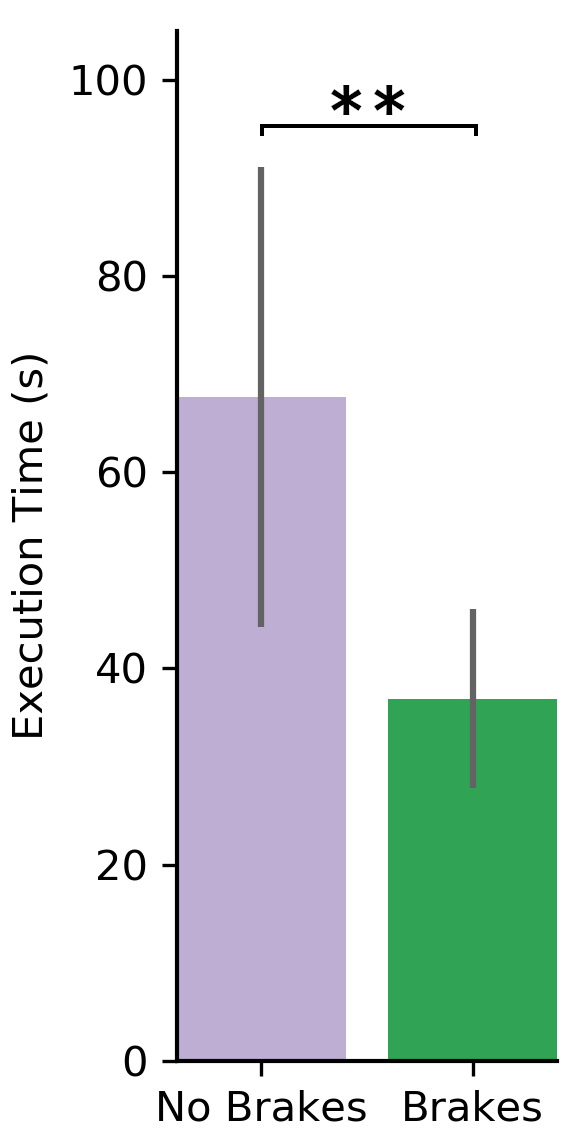}
     \end{subfigure}
     \hfill
     \begin{subfigure}[b]{0.45\columnwidth}
         \centering
         \includegraphics[width=\textwidth]{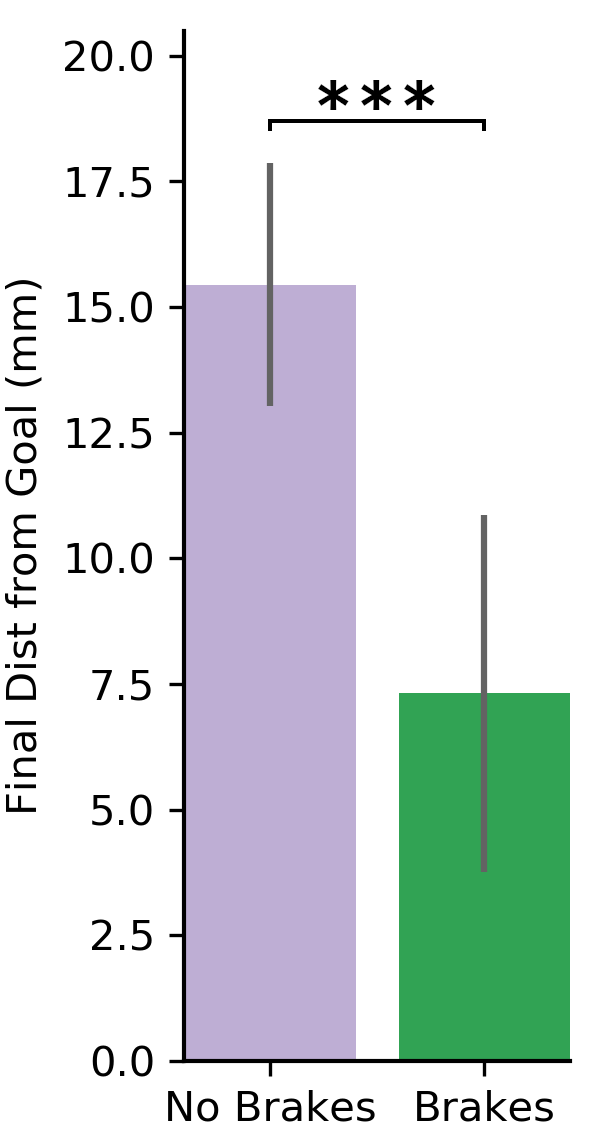}
     \end{subfigure}
        \caption{Comparison of in-hand manipulation performance when using or not using brakes. Error bars correspond to one standard deviation. Increasing number of stars indicates higher significance levels ($p < 0.05$, $p < 0.01$, $p < 0.001$) according to a paired U test. (Left) The amount of time required to complete the manipulation. (Right) The distance between the final pose of the object and the desired goal pose. }
        \label{fig:hand_result}
\end{figure}

\subsubsection{Results}

We observe that the use of electrostatic brakes significantly improves the dexterity of the robot hand. Without the brakes, the system is limited to a space of trajectories that correspond to just pulling and releasing the tendons. The flexibility to control individual joints via braking allows the controller to find trajectories that are both faster and terminate more closely to the goal pose. 

The robot hand was only able to complete nine out of ten manipulation trials without the use of brakes,  but succeeded in all ten manipulation trials that did use brakes. During the brakeless manipulation trial that failed, the robot-object system reached a state in which the controller could not find actions that would bring the object closer to the goal position. We considered this trial to be a failure after the robot was unable to make any progress towards the goal over the course of several minutes. Note that all of the following statistics are computed using only the successful manipulation trials. We found that the use of brakes significantly reduces the amount of time required to complete the in-hand manipulation task ($p < 0.01$, Mann-Whitney U test) as shown in Fig. \ref{fig:hand_result}. The robot was able to complete the task in 36.9 seconds on average when using brakes, which is 45\% faster than the 67.7 second average execution time observed when not using brakes. Upon completing the task, brake-aided manipulation had positioned the object significantly closer to the goal position compared to manipulation without brakes ($p < 0.001$, U test). Without brakes, the robot achieved an average final positioning error of 15.4 mm. Brake-aided manipulation achieved 7.3 mm final positioning error on average, which represents a 53\% reduction in error relative to manipulation without brakes.

\section{Discussion}

Our experiments demonstrate that electrostatic braking is an attractive actuation strategy whenever full actuation of a multi-jointed mechanism violates weight, volume, or power consumption constraints, but conventional underactuation does not provide sufficient control. Stacking thin brakes on top of each other allows for braking capability to be scaled as necessary regardless of whether the corresponding joint is large or small. We proposed a joint design that optimizes electrode conformance by converting rotational motion of the joint into linear sliding between the electrodes. We observed that even with multiple stacks of brakes, the joint design achieves optimal electrode conformance. Specifically, measurements of the joint's braking strength were as large as that predicted by its theoretical model. In comparison to motors, our brake-equipped joint is four times lighter and one thousand times more power efficient while exerting
similar holding torques. 

We use this joint design to develop two different underactuated, highly articulated robots. First, we built a ten degree-of-freedom robot that uses a single motor in concert with its brakes to cage and manipulate multiple objects. A conventionally underactuated version of this robot would not be able to complete this task. Second, we built a brake-equipped two-fingered robot for in-hand manipulation. For a task that required the robot to move an object from one side of its workspace to the other, we found that the use of brakes allowed the robot to complete the task 45\% faster and with 53\% less error in final positioning accuracy than without brakes. We demonstrated brake-aided actuation for two highly articulated robots, but it is applicable to any mechanism that uses tendons to couple together a set of two or more joints. In order to gain even greater control over the motion of the joints, future electrostatic brakes could exert a continuum of braking forces (rather than only being on or off) by low-pass filtering a pulse-width modulated voltage. 

\subsection{Limitations}

Unfortunately, the use of relatively high voltages can be an impediment to the wide spread adoption of electrostatic brakes due to the need for additional safety precautions, high voltage tolerant electronics, and electric spark resistant fabrication. While stacking of electrodes allows braking capability to be increased linearly, it will unlikely be able to compensate for the quadratic reduction in braking capability accompanied with lowering the applied voltage. Instead, the development of insulators that are increasingly thin yet characterized by low cost, mechanical ruggedness, and high permittivity will allow the voltage to be reduced and, more generally, encourage further adoption of electrostatic braking in robot joints. \rev{At a systems level, we initially chose a conservative movement speed for our robots and did not explore increasing it. While there is evidence in other works that individual electrostatic brakes can engage and disengage on the order of tens of milliseconds ~\citep{hinchet2020high}, it is future work to explore the maximum speeds at which electrostatic brake enabled robotic mechanisms can operate. Although this work demonstrated the ability of hybrid motor-brake mechanisms to do two different manipulation tasks, this actuation strategy is fundamentally underactuated. Future designers of robots should weigh the benefits and costs of full actuation solely with electromechanical motors versus the reduced size, weight, and power consumption of hybrid motor-brake actuation.}

\subsection{Implications}

The dexterity of conventional robots is typically dependent on the use of many motors. Replacing conventional motors with electrostatic brakes is a design direction that enables significant reductions in a robot's cost, weight, volume, and power consumption while maintaining the ability to reach arbitrary joint configurations. This approach will enable robots to achieve dexterous movement in scenarios that were not previously possible. 

Practical mobile manipulation tasks often require the robot to be untethered from remote power sources. Instead, the robot will typically draw energy from on board batteries that implicitly dictate the amount of time that the robot can operate. Weight will also influence operation time as heavier robots will need to expend more energy to move and/or manipulate. Autonomous aerial vehicles (AAV) equipped with manipulators exemplify such a scenario~\citep{baizid2017behavioral,orsag2014valve}. While AAVs can be deployed to areas that are inaccessible to land-based vehicles, their battery life is typically on the order of tens of minutes. The use of manipulators further decreases the battery life of an AAV, but the development of light weight, power efficient brake-aided manipulators will mitigate the amount of energy spent on manipulation. Mobile manipulation is just one area in which electrostatic braking can address the need for robots to perform dexterous movement in weight, volume, or power constrainted settings; others include prosthetics, climbing robots, and undersea robots.

Although the robots from this work have rigid structures, electrostatic brakes are also well suited for installation into soft robots due to the flexibility of their underlying materials~\citep{polygerinos2017soft,manti2016stiffening}. In particular, thin electrostatic brakes could be embedded in the skin of biologically inspired robots to either help actuate the joints of an internal skeleton or dynamically create joints in completely soft robots. Skin-embedded electrostatic brakes could serve a dual purpose; not only providing a means of actuation, but also leveraging the capacitive nature of the brake to sense external forces applied to the skin. 




\begin{dci}
Authors declare that they have no competing interests.
\end{dci}

\begin{acks}
The authors thank Pratik Gyawali for help with modeling and control of the robot hand in simulation. This work was supported by the National Science Foundation (EFRI grant 1832795, IIS grant 2007011, DMS grant 1839371), the Office of Naval Research, the US Army Research Laboratory CCDC, Amazon, and the Honda Research Institute USA.
\end{acks}

\begin{das}
All data are available in the main text or the supplementary materials.
\end{das}

\balance
\bibliographystyle{SageH}
\bibliography{brakes_bib}





\end{document}